\documentclass{article}
\usepackage[square,numbers,compress]{natbib}
\usepackage{times}
\usepackage{latexsym}

\usepackage[T1]{fontenc}

\usepackage[utf8]{inputenc}

\usepackage{microtype}

\usepackage{inconsolata}

\usepackage{graphicx}
\usepackage{hyperref}       
\usepackage{url}            
\usepackage{booktabs}       
\usepackage{amssymb,amsfonts,amsmath}
\usepackage{nicefrac}       
\usepackage{microtype}      
\usepackage[table,dvipsnames]{xcolor}         
\usepackage{tabularx}
\usepackage{booktabs}
\usepackage{multirow} 
\usepackage{array}
\usepackage{arydshln}
\usepackage{color, colortbl}
\usepackage{subcaption}
\usepackage{tabu}
\usepackage{pifont}
\usepackage{makecell}
\usepackage[most]{tcolorbox}
\tcbuselibrary{listings}

\newenvironment{promptbox}[1]{%
  \tcblisting{
    title=#1,
    colback=gray!5,
    colframe=black!80,
    boxrule=0.3pt,
    arc=1pt,
    left=2pt,
    right=2pt,
    top=2pt,
    bottom=2pt,
    breakable,
    listing only,
    listing options={
      basicstyle=\ttfamily\footnotesize,
      breaklines=true,
      columns=fullflexible,
      keepspaces=true
    }
  }
}{%
  \endtcblisting
}

\usepackage{lipsum}  
\usepackage{paralist}
\usepackage{threeparttable}
\usepackage{hhline}
\usepackage{scalerel,xparse}
\usepackage{appendix}
\usepackage{float}
\usepackage[table]{xcolor}
\usepackage{longtable}
\usepackage{colortbl}
\usepackage{pdflscape}
\usepackage{pgfplotstable}
\pgfplotsset{compat=1.17}
\usepackage{caption}
\usepackage{soul} 

\usepackage{xspace}

\usepackage[shortlabels]{enumitem}
\setlist[itemize]{leftmargin=*}
\setlist[enumerate]{leftmargin=*}
\usepackage[normalem]{ulem}
\usepackage{todonotes}

\newcommand{\simulationname}{SWR} 

\definecolor{lightblue}{HTML}{C7E0EB} 
\definecolor{deepblue}{HTML}{82B9D3}  
\definecolor{midblue}{HTML}{7E9CD2}   

\definecolor{lightred}{HTML}{F0C8C6}  
\definecolor{midred}{HTML}{EAB0AB}    
\definecolor{darkred}{HTML}{DB776F}   

\usepackage[final]{neurips_2025}

\usepackage[utf8]{inputenc} 
\usepackage[T1]{fontenc}    
\usepackage{hyperref}       
\usepackage{url}            
\usepackage{booktabs}       
\usepackage{amsfonts}       
\usepackage{nicefrac}       
\usepackage{microtype}      
\usepackage{xcolor}         
\usepackage{xpatch}
\usepackage{graphicx}
\usepackage{float}
\usepackage{caption}
\usepackage{booktabs}
\usepackage{multirow}
\usepackage{pifont} 
\usepackage[table]{xcolor}
\usepackage[linesnumbered,ruled,vlined]{algorithm2e}
\definecolor{headerblue}{RGB}{242,248,254}
\definecolor{myPink}{RGB}{228,85,160}
\usepackage{amsmath}
\usepackage{enumitem}

\makeatletter
\xapptocmd{\NAT@bibsetnum}{\setlength{\leftmargin}{0pt}\setlength{\itemindent}{\labelwidth}\addtolength{\itemindent}{\labelsep}}{}{}
\makeatother

\title{SimWorld-Robotics: Synthesizing Photorealistic and Dynamic Urban Environments for Multimodal Robot Navigation and Collaboration}

\newcommand{\uva}{$^{1}$}
\newcommand{\ucsd}{$^{2}$}
\newcommand{\jhu}{$^{3}$}
\newcommand{\umich}{$^{5}$}
\newcommand{\cmu}{$^{4}$}
\newcommand{\colead}{$^{*}$}
\newcommand{\coadvise}{$^{\ddag}$}

\author{%
    Yan Zhuang\uva\hspace{6pt}
    Jiawei Ren\ucsd\colead\hspace{6pt}
    Xiaokang Ye\ucsd\colead\hspace{6pt}
    Jianzhi Shen\jhu\hspace{6pt}
    Ruixuan Zhang\jhu\hspace{6pt} \\
    {\bf
    Tianai Yue\jhu\hspace{6pt}
    Muhammad Faayez\jhu\hspace{6pt}
    Xuhong He\cmu\hspace{6pt}
    Xiyan Zhang\jhu}\\
    {\bf
    Ziqiao Ma\umich\hspace{6pt}
    Lianhui Qin\ucsd\coadvise\hspace{6pt}
    Zhiting Hu\ucsd\coadvise\hspace{6pt}
    Tianmin Shu\jhu\coadvise}\\
    \uva University of Virginia\;\hspace{6pt}
    \ucsd UC San Diego\;\hspace{6pt}
    \jhu Johns Hopkins University\; \\
    \cmu Carnegie Mellon University\; \hspace{6pt}
    \umich University of Michigan\; 
}

\begin{document}

\maketitle

\let\thefootnote\relax\footnotetext{*~Equal contribution. \ddag~Equal advising.}

\begin{abstract}
    Recent advances in foundation models have shown promising results in developing generalist robotics that can perform diverse tasks in open-ended scenarios given multimodal inputs. However, current work has been mainly focused on indoor, household scenarios. In this work, we present SimWorld-Robotics~(\simulationname), a simulation platform for embodied AI in large-scale, photorealistic urban environments. Built on Unreal Engine 5, \simulationname\ procedurally generates unlimited photorealistic urban scenes populated with dynamic elements such as pedestrians and traffic systems, surpassing prior urban simulations in realism, complexity, and scalability. It also supports multi-robot control and communication. With these key features, we build two challenging robot benchmarks: (1) a multimodal instruction-following task, where a robot must follow vision-language navigation instructions to reach a destination in the presence of pedestrians and traffic; and (2) a multi-agent search task, where two robots must communicate to cooperatively locate and meet each other. Unlike existing benchmarks, these two new benchmarks comprehensively evaluate a wide range of critical robot capacities in realistic scenarios, including (1) multimodal instructions grounding, (2) 3D spatial reasoning in large environments, (3) safe, long-range navigation with people and traffic, (4) multi-robot collaboration, and (5) grounded communication. Our experimental results demonstrate that state-of-the-art models, including vision-language models (VLMs), struggle with our tasks, lacking robust perception, reasoning, and planning abilities necessary for urban environments. 
    


\end{abstract}

\vspace{-15pt}
\begin{center}
\texttt{Project website: \href{https://scai.cs.jhu.edu/projects/SimWorldRobotics/}{SimWorld-Robotics}}
\end{center}
\vspace{-15pt}

\section{Introduction}
\begin{figure}[!t]
    \centering
    \includegraphics[trim=55 90 48 90, clip, width=\linewidth]{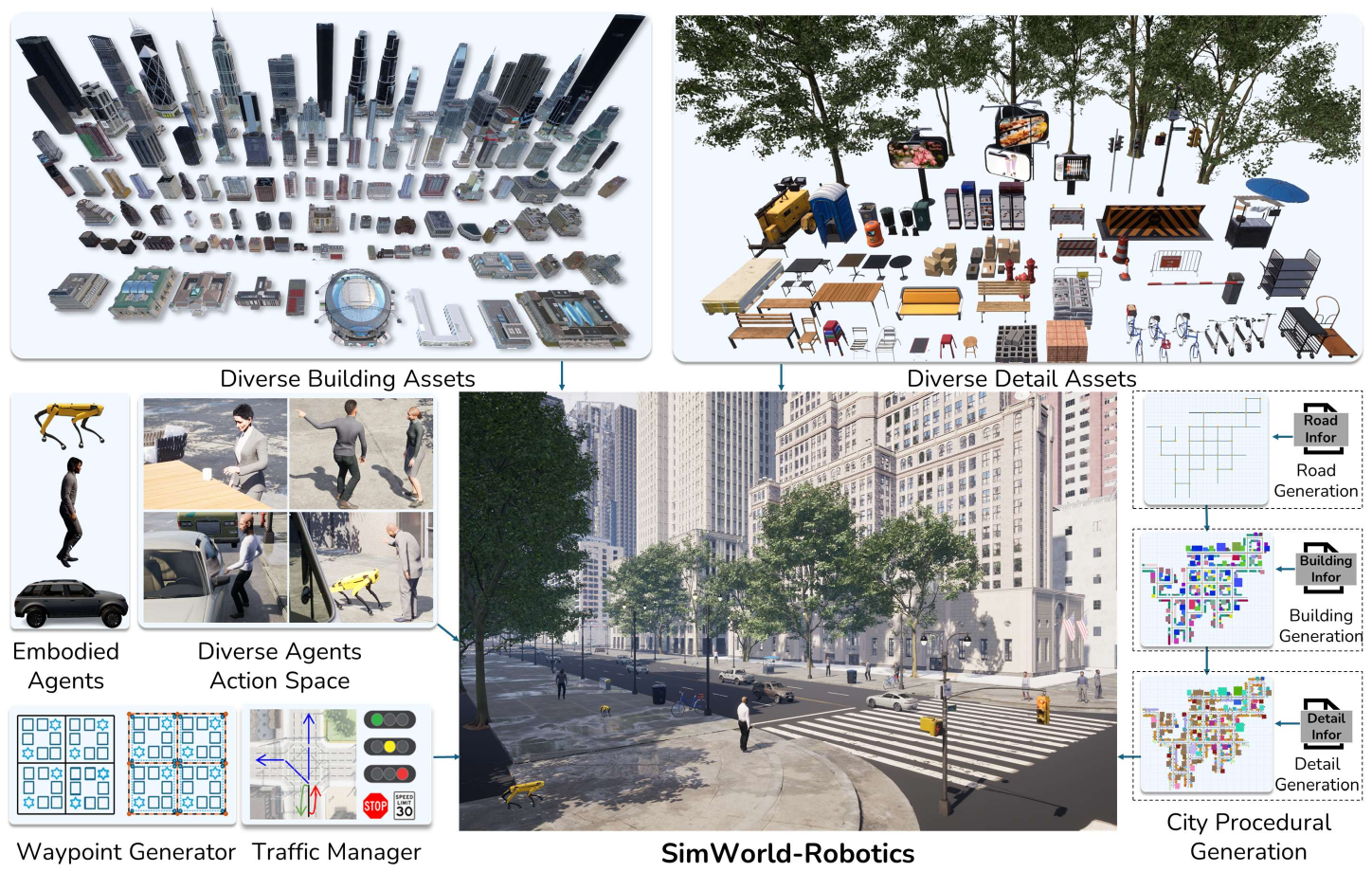}
    \caption{\textbf{Overview of SimWorld Robotics~(\simulationname).} Built upon Unreal Engine 5, \simulationname\ is a simulation platform for large-scale, photorealistic, and dynamic urban environments. It offers diverse high-fidelity building and object assets, supports embodied agents with rich action spaces, includes a background traffic system powered by city-scale waypoint generation, and enables comprehensive city procedural generation.} 
    \vspace*{-20pt}
    \label{fig:overview} 
\end{figure}

There has been tremendous progress in engineering general-purpose robotics that can follow human instructions and perform open-ended tasks~\cite{anderson2018vision, NEURIPS2024_d8087dea, gan2021threedworldtransportchallengevisually, pmlr-v205-li23a, Shridhar_2020_CVPR}, thanks to the advances in robot foundation models. Training these models requires a large amount of data, much of which can be generated in high-fidelity embodied simulators, such as Habitat 3~\cite{puig2023habitat30cohabitathumans}, RoboTHOR~\cite{deitke2020robothoropensimulationtorealembodied}, TDW~\cite{gan2021threedworldplatforminteractivemultimodal}, VirtualHome~\cite{puig2018virtualhomesimulatinghouseholdactivities}, Virtual Community~\cite{zhou2025virtualcommunityopenworld} and BEHAVIOR~\cite{pmlr-v205-li23a}. They can also be systematically evaluated in diverse scenarios created in these simulators. However, current embodied simulators for robotics have been focused on tabletop~\cite{mu2021maniskillgeneralizablemanipulationskill, yin2025partinstructpartlevelinstructionfollowing, mees2022calvinbenchmarklanguageconditionedpolicy, jiang2023vimageneralrobotmanipulation, zhang2023lohoravenslonghorizonlanguageconditionedbenchmark} or household tasks~\cite{szot2022habitat20traininghome, pmlr-v205-li23a, li2021igibson20objectcentricsimulation, srivastava2021behaviorbenchmarkeverydayhousehold, Shridhar_2020_CVPR}. In this work, we want to study how to create a realistic and scalable embodied simulator for outdoor robotics tasks.

Compared to indoor scenarios, robotics in outdoor environments, in particular, large urban environments, introduces additional challenges, such as (1) 3D perception, spatial reasoning and grounding in large environments; (2) safe navigation in dynamic scenes with people and traffic; (3) long-range spatial memory; and (4) multi-agent collaboration and communication in task.

\begin{figure}[!t]
    \centering
    \includegraphics[trim=0 215 -3 95, width=\linewidth]{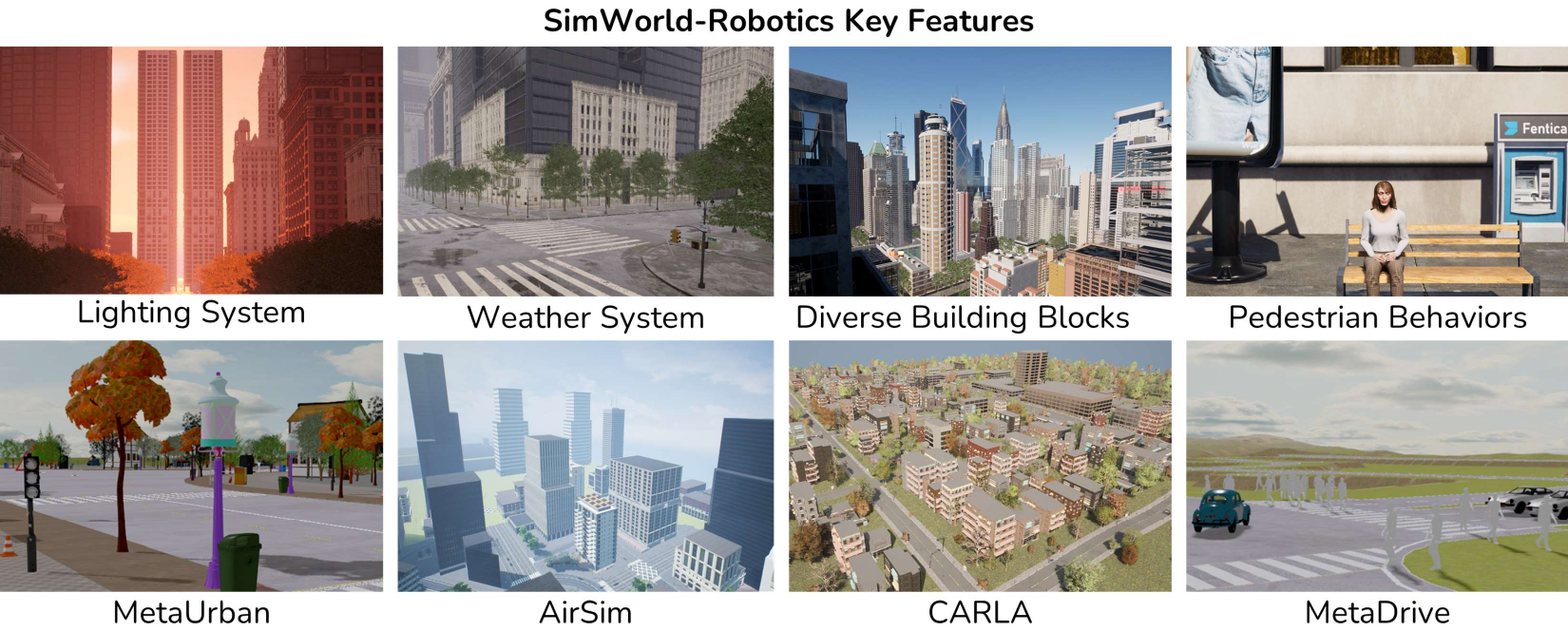}
    \caption{\textbf{Simulator Comparison} Top: Our simulator demonstrates key features including dynamic lighting (e.g., sunrise), realistic weather (e.g., rain), diverse high-fidelity buildings, and rich pedestrian behaviors. Bottom: MetaUrban and MetaDrive support large-scale maps but suffer from poor rendering quality; AirSim and CARLA offer better visuals but have limited building diversity and high repetition. All existing simulators exhibit simplistic pedestrian behaviors limited to random walking, failing to reflect real urban dynamics.}
    \vspace*{-15pt}
    \label{fig:simulator_comparison} 
\end{figure}

There have been urban simulators developed in recent years. However, to address the critical challenges faced by real-world robotics in urban environments, they lack the necessary realism, customizability, scalability, and versatility. For instance, well-known simulators such as AirSim~\cite{shah2017airsimhighfidelityvisualphysical}, CARLA~\cite{Dosovitskiy17} mainly focus on autonomous driving domains. While it supports some manual customization to the provided city environments, it does not support procedural city environment generation. It also does not support the flexible control of embodied agents (such as mobile robots or pedestrians) other than vehicles. More recent city simulators, such as MetaDrive~\cite{li2022metadrivecomposingdiversedriving}, MetaUrban~\cite {wu2024metaurbanembodiedaisimulation}, significantly improve the scalability. However, the simulated environments still lack photorealism as shown in Figure~\ref{fig:simulator_comparison}.

Therefore, we introduce SimWorld-Robotics~(\simulationname), a new embodied AI simulation platform for large-scale, photorealistic, and dynamic urban environments. As illustrated in Figure~\ref{fig:overview}, \simulationname{} offers diverse high-fidelity building and object assets, multiple types of embodied agents with rich action spaces, a waypoint-based background traffic system, and a comprehensive procedural city generation pipeline to generate infinite cities. \simulationname{} also supports scalable data generation with fine-grained ground-truth annotations, enabling the training and evaluation of embodied agents at scale. Together, these features accelerate progress toward stronger embodied intelligence.

\begin{figure}[t!]
    \centering
    \includegraphics[trim=5 250 0 88, clip,width=\linewidth]{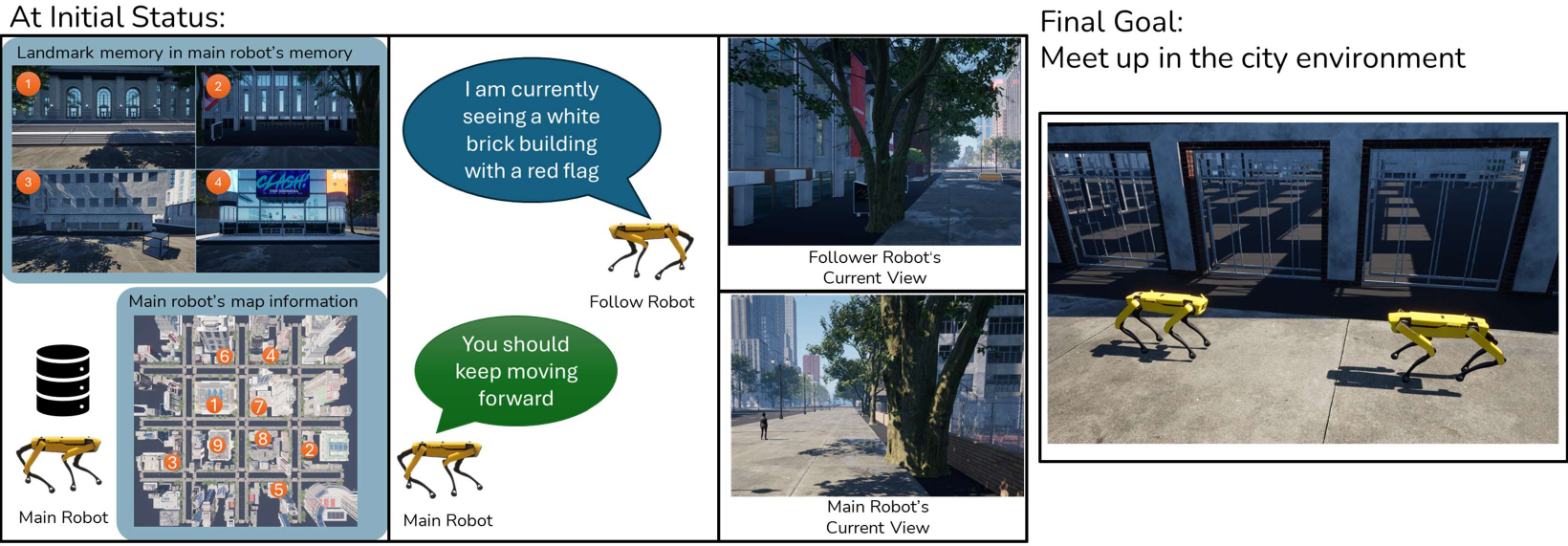} 
    \vspace*{-10pt}
    \caption{Illustration of a multi-robot search task.} 
    \vspace*{-10pt}
    \label{fig:multi-agent task} 
\end{figure}

By leveraging \simulationname{}, we develop two novel benchmarks for robots in large, urban environments. Each benchmark evaluates crucial robot capabilities uniquely supported by the key features of \simulationname{}. As shown in~\ref{fig:single-agent task}, the first is a multimodal instruction following benchmark, \textsc{SimWorld-MMNav}, for robot navigation, in which a robot must follow vision and language instructions to reach the target location. Unlike existing robot navigation benchmarks, we evaluate multiple robot capacities necessary for real-world urban navigation jointly, including robust 3D visual perception, grounding multimodal instructions to 3D environments, obstacle avoidance, following traffic rules, and walking around people in a socially acceptable way. The second is a multi-robot search benchmark, \textsc{SimWorld-MRS}, in which two robots must cooperate to localize and meet each other via physical navigation and verbal communication illustrated in~\ref{fig:multi-agent task}. To be successful in this benchmark, robots must be able to effectively communicate about each other's locations in a large, unfamiliar space and discuss a joint plan, combined with their 3D spatial reasoning abilities. Our experimental results demonstrate that existing models, including state-of-the-art vision-language models (VLMs), fail to achieve meaningful success on our benchmarks. This highlights the gap in current foundation models for challenging, realistic robot tasks in urban environments. 

To address this gap, we introduce SimWorld-20K, a large-scale dataset for benchmarking multimodal robot navigation in photo-realistic urban environments. The dataset contains 20K training steps sampled from 200 episodes, each averaging $500\,\mathrm{m}$ in length, across 100 procedurally generated city environments with an average area of $2\,\mathrm{km}^2$. Compared to MetaUrban~\cite{wu2024metaurbanembodiedaisimulation}, the most recent urban simulator supporting procedural city generation, \simulationname{} offers environments that are 100$\times$ larger in area and episodes that are over 1.2$\times$ longer (MetaUrban: 410 meters per episode and $0.02\,\mathrm{km}^2$ on average). This enables evaluation of long-horizon, real-world-scale navigation. After fine-tuning on SimWorld-20K, QwenVL2.5-7B achieves a non-zero success rate on the test set and outperforms SOTA proprietary models across several key metrics.

In sum, our key contributions include: (1) a new embodied AI simulator, SimWorld-Robotics~(\simulationname), that supports the creation and simulation of photorealistic and dynamic urban environments with diverse embodied agents; (2) two novel benchmarks for single robot navigation and multi-robot search tasks that evaluate key robot capacities by leverage the key, unique features of our simulator; (3) a large-scale training dataset, SimWorld-20K, that enables long-horizon multimodal robot navigation across city-scale environments; (4) a systematic evaluation of recent baseline models which identifies the significant limits of these models on the evaluated key capacities.

\section{Related Work}

\textbf{Embodied Simulators for Urban Environments.} Recent embodied simulators mostly focus on the indoor environment, mainly designed for household robots~\cite{chang2017matterport3dlearningrgbddata,tsoi2022sean2,puig2018virtualhomesimulatinghouseholdactivities, puig2021watchandhelpchallengesocialperception,gan2021threedworldplatforminteractivemultimodal, puig2023habitat30cohabitathumans, kolve2022ai2thorinteractive3denvironment,pmlr-v205-li23a}. There have been outdoor simulators \cite{Dosovitskiy17, shah2017airsimhighfidelityvisualphysical, wu2024metaurbanembodiedaisimulation, gao2024embodiedcitybenchmarkplatformembodied,wang2024grutopiadreamgeneralrobots,fan2023aerialvisionanddialognavigation} that provide more open-ended tasks, facilitate more complicated dynamics, but face challenges in reality and multi-agent parallelism. CARLA\cite{Dosovitskiy17} is a flexible simulator dedicated to autonomous driving. AirSim~\cite{shah2017airsimhighfidelityvisualphysical} focuses on simulation for drone control. EmbodiedCity~\cite{gao2024embodiedcitybenchmarkplatformembodied} provides an arena for embodied AI with realistic urban settings and enriched agent control. Grutopia~\cite{wang2024grutopiadreamgeneralrobots} offers a large-scale annotated scene dataset for general service robot tasks. AerialVLN~\cite{liu2023aerialvlnvisionandlanguagenavigationuavs} serves as a highly photorealistic urban landscape for UAV navigation. MetaUrban~\cite{wu2024metaurbanembodiedaisimulation} features a wide range of micromobility tasks on the pedestrian, creating a highly complex urban environment. However, the dynamic scene interaction of human agents and social navigation (e.g., traffic rule following and pedestrian avoidance) has been much less explored.

Our simulator places the robot in a photorealistic, dynamically populated urban environment, evaluating not only multimodal instruction following but also the robot’s ability to adapt to real-time social cues and navigate amidst human agents in a safe, rule-aware manner.


\textbf{Instruction Following Benchmarks for Robot Navigation.} Following complex instructions in the real world often requires understanding both language and vision cues in tandem. As summarized in Table~\ref{tab:benchmark_comparison} in Appendix~\ref{sec:compare_nav}, prior benchmarks on vision-and-language navigation have tackled instruction following, but typically in static indoor scenes or street panoramas using only textual descriptions\cite{pérezdarpino2020robotnavigationconstrainedpedestrian}. For instance, Touchdown~\cite{chen2020touchdownnaturallanguagenavigation} explores city navigation via language alone in a static street view setting, and R2R~\cite{anderson2018vision} focuses on multi-step instructions in household tasks. These settings lack moving obstacles and additional visual guidance, making them only partial proxies for real-world urban navigation. In contrast, our multimodal robot navigation benchmark, \textsc{SimWorld-MMNav}, challenges a robot to interpret and follow multimodal instructions (paired language instructions and visual hints) to reach the target location in a large-scale, photorealistic, dynamic city environment.





\textbf{Multi-Robot Collaboration Benchmarks.}
There have been many recent multi-robot collaboration benchmarks as summarized in Table~\ref{tab:multiagent_benchmark_comparison}, but they do not evaluate multi-robot collaboration and communication for exploration and navigation in large urban environments. RoCo~\cite{mandi2023rocodialecticmultirobotcollaboration} studies cooperative manipulation on tabletop tasks, without the need for environmental exploration. 
DOROTHIE~\cite{ma2022dorothiespokendialoguehandling} introduces spoken dialogue for an ego vehicle, yet involves just one controllable agent. 
DriVLMe~\cite{huang2024drivlmeenhancingllmbasedautonomous} simulates city-scale traffic with many vehicles, but lacks any verbal communication between robots. 
RobotSlang~\cite{banerjee2020robotslangbenchmarkdialogguidedrobot} evaluates robot communication but is restricted to small environments.
\emph{Where Are You?}~\cite{hahn2021youlocalizationembodieddialog} frames localization as a two-party dialogue, but the robots do not physically navigate to meet each other through a large environment.


\begin{table}[t!]
    \begin{threeparttable}
    \centering
    \scriptsize
    \setlength{\tabcolsep}{5.2pt} 
    \renewcommand{\arraystretch}{1.2} 
    \caption{Comparison of outdoor simulation platforms across key features. The \textbf{Scenes} section includessupport for \text{Procedural Generation} ($\checkmark$: supported, $\times$: not supported), and level of \text{Photorealism}. The \textbf{Human} section summarizes scene interaction capabilities (\text{Scene-Interact.}) and the number of supported human \text{Actions}. \textbf{Embodied Agents} indicate whether robots (Rob.), humans (Hum.), and vehicles (Veh.) are supported. \textbf{Multi-agent} denotes support for asynchronous multi-agent control. Full comparisons with other types of simulators are in Table~\ref{tab:simulators_comparison_full} in Appendix~\ref{sec:app_compare_simulator}.}

    \begin{tabularx}{\textwidth}{lccccccccc@{}}
        \toprule
        \multirow{2}{*}{Simulator} & \multicolumn{2}{c}{Scenes} & \multicolumn{2}{c}{Human} & \multicolumn{3}{c}{Embodied Agents} & \multirow{2}{*}{Multi-agent} \\ 
        \cmidrule(lr){2-3} \cmidrule(lr){4-5} \cmidrule(lr){6-8} 
         & Procedural Generation & Photorealistic & Scene-Interact. & Actions & Rob. & Hum. & Veh. \\ 
        \midrule
        CARLA~\cite{Dosovitskiy17} & $\times$ & $\checkmark$ & $\times$  & 2 & $\times$  & $\checkmark$  &  $\checkmark$  &  $\checkmark$  \\ 
        AirSim~\cite{shah2017airsimhighfidelityvisualphysical} & $\times$ & $\checkmark$ & $\times$  & 2 & $\times$  & $\checkmark$  &  $\checkmark$  &  $\checkmark$ \\ 
        MetaUrban~\cite{wu2024metaurbanembodiedaisimulation}       & $\checkmark$ & $\times$ & $\times$  & 2 & $\checkmark$  & $\times$  & $\times$  & $\checkmark$  \\ 
        EmbodiedCity~\cite{gao2024embodiedcitybenchmarkplatformembodied}    & $\times$ & $\times$ & $\times$  & 2 & $\checkmark$  & $\times$  & $\times$  &  $\times$  \\ 
        Grutopia~\cite{wang2024grutopiadreamgeneralrobots}        & $\times$ & $\checkmark$ & $\times$  & $\times$ & $\checkmark$  & $\times$  & $\times$  & $\times$  \\ 
        AerialVLN~\cite{liu2023aerialvlnvisionandlanguagenavigationuavs}       & $\times$ & $\checkmark$ & $\times$  & $\times$ & $\checkmark$  & $\times$  & $\times$ & $\times$  \\   
        UnrealZoo~\cite{zhong2025unrealzooenrichingphotorealisticvirtual} & $\times$  & $\checkmark$ & $\checkmark$  & 10  & $\checkmark$  & $\checkmark$ & $\checkmark$  & $\checkmark$   \\  
        \textbf{\simulationname{} (Ours)}  & $\checkmark$ & $\checkmark$ & $\checkmark$  & 26 & $\checkmark$  & $\checkmark$  & $\checkmark$  & $\checkmark$  \\ 
        \bottomrule
    \end{tabularx}%
    \label{tab:simulators_comparison} 
    \end{threeparttable}
    \vspace{-10pt}
\end{table}

\section{SimWorld-Robotics}
Built upon SimWorld~\cite{ren2025simworldopenendedrealisticsimulator}, SimWorld-Robotics~(\simulationname) introduces key extensions including procedural city generation, a traffic system, and support for an additional embodied agent: the quadruped robot. We begin by describing how \simulationname\ procedurally generates diverse and scalable urban environments with varying specifications. We then introduce the embodied agents supported in these environments—vehicles, humans, and robots—and explain the logic behind asynchronously controlling multiple agents. Finally, we outline the rule-based system governing background pedestrians and traffic dynamics. Details of \simulationname\ can be found in Appendix~\ref{sec:app_simulator}.

\begin{figure}[t!]
    \centering
    \includegraphics[trim=10 95 10 95, clip, width=\linewidth]{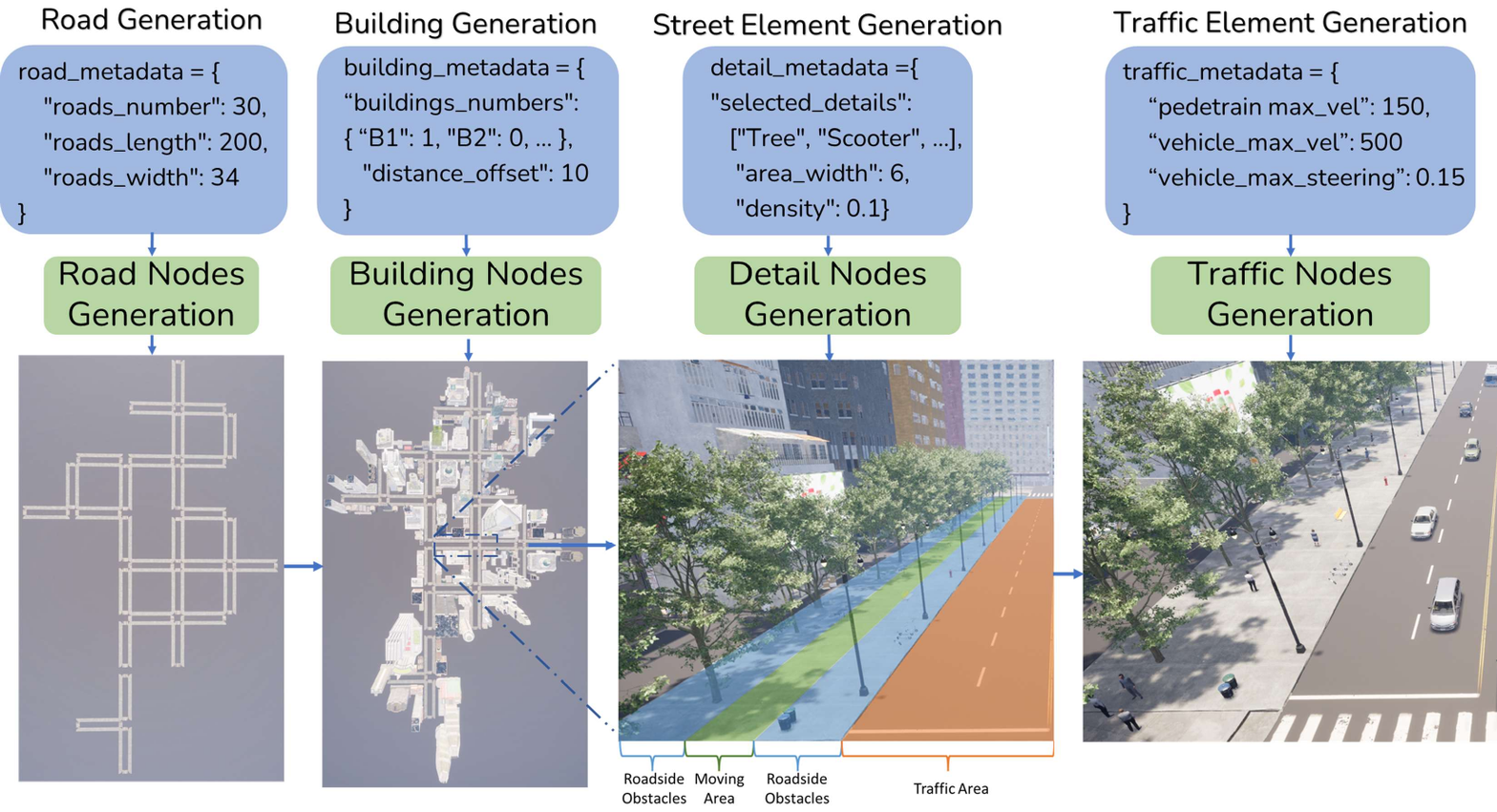}
     \vspace*{-10pt}
    \caption{\textbf{Procedural City Generation.} \simulationname\ receives a user's specification and modularizes the process into road, building, details, and traffic elements generation.} 
    \vspace*{-20pt}
    \label{fig:procedural generation} 
\end{figure}

\subsection{Procedural City Generation}

The Procedural City Generation pipeline in \simulationname\ is designed to synthesize realistic, structured urban environments from minimal specifications, supporting tasks such as autonomous driving, pedestrian navigation, and multi-agent simulations. As shown in Figure~\ref{fig:procedural generation}, the pipeline follows four stages: road, building, street element, and traffic element generation. It begins by creating a road network through a priority queue-based growth strategy that balances branching and depth, while ensuring plausible layouts via road-end attachment and intersection validation. Buildings are then placed along roads using collision-aware sampling and greedy gap-filling to maximize coverage and maintain uniformity. Next, contextual street elements—trees, cones, benches, and parked vehicles—are placed around buildings and sidewalks with basic accessibility constraints. Finally, dynamic traffic elements---including vehicles and pedestrians---are integrated into the environment to support research on traffic dynamics, agent-based behavior modeling, and realistic navigation scenarios. A detailed description of the city procedural generation pipeline is provided in Appendix~\ref{sec:app_city_generation}.

\subsection{Embodied Agents}

\simulationname\ supports three types of embodied agents—humans, vehicles, and robots. Unlike synchronous designs~\cite{puig2018virtualhomesimulatinghouseholdactivities} where agents must wait for others to complete their actions, \simulationname\ allows asynchronous control, enabling each agent to act independently. To better reflect real-world conditions and support diverse tasks, \simulationname\ also provides rich action spaces for different agents and flexible observation spaces. The creation of new embodied agents in Appendix~\ref{sec:app_customization}.

\textbf{Types of Embodied Agents.} Unlike previous simulators~\cite{Dosovitskiy17, wu2024metaurbanembodiedaisimulation, li2022metadrivecomposingdiversedriving}that focus on a specific type of embodied agent---such as autonomous vehicles, robots, or humans---and mainly support control over that particular agent type, our simulator simultaneously supports all three major categories. This unified design enables the development of broader and more diverse embodied AI tasks within a single environment. In \simulationname{}, we have included all three types of embodied agents. For the robots, we have two kinds: the scooter and the quadruped robot. 

\textbf{Asynchronous Multi-agent Control.} To realistically model scenarios where multiple agents act independently and simultaneously, \simulationname\ uses an asynchronous multi-agent control framework. Each agent receives its observation from a centralized buffer and can submit an action only when marked as available. The buffer updates at a fixed interval (default: 0.01 seconds), checking for actions from available agents and updating their availability status. Valid actions are executed concurrently, after which all agents become unavailable until their actions complete. Once finished, agents are marked as available again and receive updated observations. The control pipeline is illustrated in Figure~\ref{fig:aynchronized multi-agent control} in Appendix~\ref{sec:app_AMC}.


\textbf{Observation Space.} \simulationname\ provides three primary types of visual observations: RGB images, depth images, and semantic segmentation masks, as illustrated in Figure~\ref{fig:observation space} in Appendix~\ref{sec:app_obs}. In addition to these, \simulationname\ also offers ground-truth language descriptions and 3D bounding boxes for objects present in the environment.

\textbf{Action Space.} \simulationname\ supports three types of continuous vehicle control: acceleration, braking, and steering. Each action is continuous within a set range, allowing flexible control—e.g., higher acceleration leads to faster speeds, and larger steering values result in sharper turns. For robot control, actions include continuous translation (forward, backward, left, right) and free-angle rotation, enabling flexible movement and orientation, as illustrated in Figure~\ref{fig:observation space} in Appendix~\ref{sec:app_action_space}. Human agents can perform navigation actions (movement, turning) and interaction actions relevant to urban scenarios, grouped into: (1) human–object (e.g., pick up/drop off objects, sit/stand), (2) human–vehicle (e.g., drive, enter/exit, open/close trunk), and (3) human–human (e.g., wave, argue, point).




\subsection{Pedestrian and Traffic Simulation}

The traffic simulation in \simulationname\ creates dynamic urban scenarios by orchestrating vehicle and pedestrian movement across a generated city map. It supports route assignment, intersection control, and pedestrian flow simulation, running on a fixed-time update loop for consistent real-time updates. Vehicle motion is governed by a feedback-based model using a PID controller, with empirically tuned parameters for realistic acceleration, braking, and turning dynamics \cite{jain2024comprehensivesurveypidpure}. Pedestrians follow a lightweight model, adjusting orientation incrementally toward their goals based on angular differences. To simulate realistic patterns, \simulationname\ uses a probabilistic routing strategy at intersections, where agents select paths based on predefined probabilities. This stochastic behavior introduces natural variability and enhances scene diversity. Details of predestrain and traffic simulation can be found in Appendix~\ref{sec:app_traffic_system}.

\section{Multimodal Robot Navigation Benchmark}

\begin{figure}[!t]
    \centering
    \includegraphics[trim=0 272 0 82, clip, width=\linewidth]{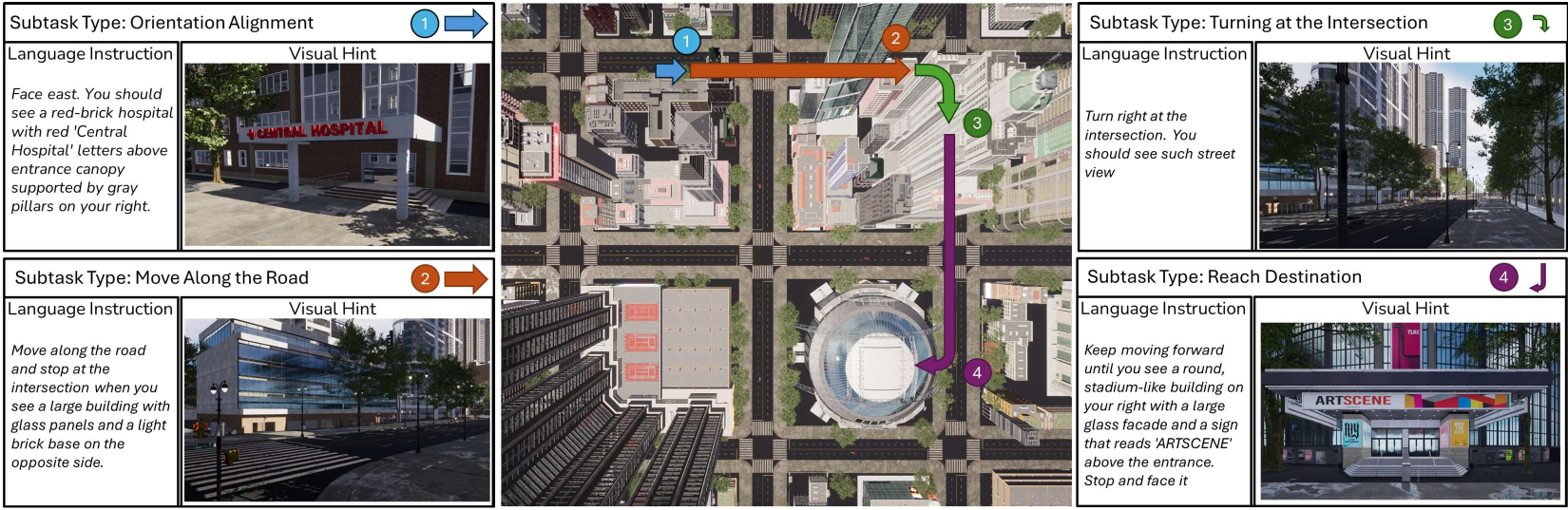} 
    \vspace*{-10 pt}
    \caption{Illustration of a multimodal robot navigation task.} 
    \vspace*{-15pt}
    \label{fig:single-agent task} 
\end{figure}

We propose a novel multimodal robot navigation benchmark, \textsc{SimWorld-MMNav}, where a robot must follow multimodal instructions—paired language and visual hints—to reach a target in a large-scale, photorealistic, dynamic city environment (Figure~\ref{fig:single-agent task}). This requires grounding verbal and visual references in the 3D environment based on the robot’s observations, while adapting to real-world complexities like traffic and pedestrians. \simulationname{} enables scalable evaluation of such tasks through (1) procedurally generated instructions with annotated, richly detailed city layouts, (2) photorealistic rendering of visual hints, and (3) simulation of pedestrians and traffic. We highlight core challenges of dynamic grounding: aligning language with visual targets, navigating around moving entities, and obeying environmental rules—all within an open-world urban setting that surpasses prior benchmarks. Details of \textsc{SimWorld-MMNav} can be found in Appendix~\ref{sec:app_MMNAv}.

\subsection{Setup}

\begin{table}[!t]
    \centering
    \caption{Experimental results on the \textsc{SimWorld-MMNav} benchmark (easy task set). The numbers in parentheses indicate the improvement after finetuning.}
    \scriptsize
    \setlength{\tabcolsep}{14pt}
    \label{tab:mmnav}
    \resizebox{\textwidth}{!}{%
    \begin{tabular}{lccc}
    \toprule
    \textbf{Models} & \textbf{SR\%$\uparrow$} & \textbf{Subtask SR\% $\uparrow$} & \textbf{Distance Progress\% $\uparrow$} \\
    \midrule
    \rowcolor{gray!10} \multicolumn{4}{l}{\textit{Proprietary Models}} \\
    GPT-4o & 0 & 33.07 & 15.60 \\
    Gemini 2.5 Flash & 0 & 37.06 & 31.29 \\
    \midrule
        \rowcolor{gray!10} \multicolumn{4}{l}{\textit{Reasoning Models}} \\
    GPT-o3 & 5.0 & 42.50 & 38.43 \\
    GPT-o3-pro & 8.3 & 46.35 & 39.46 \\
    \midrule
    \rowcolor{gray!10} \multicolumn{4}{l}{\textit{Open-sourced Models}} \\
    QwenVL 2.5 7B & 0 & 16.86 & 7.82 \\
    QwenVL 2.5 72B & 0 & 23.80 & 17.50 \\
    Gemma 3 27B & 0 & 15.36 & 6.83 \\
    InternVL 3 78B & 0 & 18.31 & 9.34 \\
    \midrule
    \rowcolor{gray!10} \multicolumn{4}{l}{\textit{Fine-tuned Models}} \\
    QwenVL 2.5 7B$_{ft}$  & 4.0 (+4.0) & 52.45 (+35.59) & 53.63 (+45.81) \\
    \midrule
    \rowcolor{gray!10} \multicolumn{4}{l}{\textit{Hybrid Baselines}} \\
    HybridGPT & 0 & 32.53 & 27.24 \\
    \midrule
    \rowcolor{gray!10} \multicolumn{4}{l}{\textit{RL-based Baselines}} \\
    VLA-RL & 0 & 28.37 & 22.79 \\
    \bottomrule
    \end{tabular}
    }
    \vspace{-10pt}
\end{table}

\textbf{Task Definition.}
Each task consists of a series of multimodal instructions $I_{1\ldots K}$ that the robot must interpret and execute. As illustrated in Figure~\ref{fig:single-agent task}, each instruction $I_k$ includes a natural language instruction and a visual hint illustrating what the robot can expect to see after reaching the goal described by the language instruction. These two modalities together guide the robot toward a specific location and/or orientation. Each instruction is associated with a ground-truth goal $g_k$. At any given timestep, only one instruction is presented to the robot, which must successfully complete the current instruction (i.e., reaching $g_k$) before receiving the next instruction $I_{k+1}$. 

In real-world urban navigation, instructions typically fall into one of four categories: Orientation Alignment, Move Along the Road, Turn at the Intersection, and Reach Destination. As shown in Figure~\ref{fig:single-agent task}, a complete sequential instruction-following task is formed by chaining together multiple different types of instructions. Details of procedural task generation can be found in Appendix~\ref{sec:app_task_generation}.

To closely mirror real-world navigation challenges, the environments simulated incorporate both static obstacles (e.g., buildings, barriers) and dynamic agents (e.g., pedestrians, vehicles) whose trajectories are not known in advance. Task difficulty is determined by the complexity of the environment, and we define two levels: an \textit{easy} task includes no obstacles (static objects outside of buildings, pedestrians or vehicles), and a \textit{hard} task adds both static object obstacles and dynamic pedestrians and vehicles.

\textbf{Obsevation and Action Space}. Apart from the subtask description and expected visual hint, the robot has access to the egocentric RGB image, segmentation image, and depth image. Ground truth orientation is provided, assuming a built-in compass in the robot. The robot's action space includes moving in four directions, as well as turning. It can also stay still and confirm task completion. \simulationname{} also supports active perception by providing different viewpoint images.

\textbf{SimWorld-20k}. Based on \simulationname{},  we construct the SimWorld-20k dataset, including 100 training maps with an average area of $2\,\mathrm{km}^2$ and 200 oracle trajectories generated by A*~\cite{Hart1968} for 200 training tasks, whose average length is greater than $2.5\,\mathrm{km}$.  Each trajectory contains over 100 steps, forming a training dataset of 20K steps. Details of it can be found in Appendix~\ref{sec:app_20k}.

\textbf{Statistics.} 
For both training and evaluation, we synthesize 100 distinct worlds. With them, we create 200 easy tasks and 200 hard tasks. Each task has 2-4 instructions. On average, each task requires traveling 500 meters over 250 steps. To ensure generalization, 33\% of the buildings in the testing set are exclusive and unseen during training. 

\textbf{Evaluation Metrics.} We evaluate navigation performance in the easy setting through three key metrics: (1) \textbf{Success Rate (SR)} \cite{anderson2018evaluationembodiednavigationagents} measures the percentage of goal arrivals, (2) \textbf{Subtask SR} \cite{Shridhar_2020_CVPR} tracks the ratio of completed subtasks, (3) \textbf{Distance Progress} \cite{thomason2019visionanddialognavigation} evaluates instruction-following by the relative reduction in distance to the goal. The hard evaluation setting, which introduces a comprehensive traffic system and pedestrians, requires three additional safety metrics: (1) \textbf{Static Collision} counts collision between the robot and immovable objects like buildings and trees, (2) \textbf{Dynamic Collision} counts collisions with moving elements, such as pedestrians and vehicles, (3) \textbf{Traffic Light Violation} records the number of times the robot fails to adhere to traffic signal regulations.


\begin{table}[!t]
    \centering
    \setlength{\tabcolsep}{1pt}
    \caption{Experimental results on the \textsc{SimWorld-MMNav} benchmark (hard task set).}
    \label{tab:single-agent-hard}
    \resizebox{\textwidth}{!}{%
    \begin{tabular}{lcccccc}
    \toprule
    \textbf{Models} & \textbf{SR\%$\uparrow$} & \textbf{Stat. Coll.
$\downarrow$} & \textbf{Dyn. Coll.$\downarrow$} & \textbf{Red Light Viol.$\downarrow$} & \textbf{Subtask SR\%$\uparrow$} & \textbf{Distance Progress\%$\uparrow$} \\
    \midrule
    GPT-4o & 2.08 & 1.92 & 10.37  & 3.02 & 34.38 & 24.83 \\
    Gemini 2.5 Flash & 0 & 3.21 & 4.29 & 7.875 & 32.29 & 29.87 \\
    QwenVL 72B & 0 & 5.0 & 11.73 & 2.86 & 23.86 & 21.97 \\
    \bottomrule
    \end{tabular}
    }
    \vspace{-20pt}
\end{table}

\begin{table}[H]                                     
\centering
\begin{threeparttable}
\caption{Most common failure modes in \textsc{SimWorld-MMNav}.}
\label{tab:failure_modes}
\footnotesize
\setlength{\tabcolsep}{0pt}
\begin{tabular*}{\textwidth}{@{\extracolsep{\fill}}llc@{}}
\toprule
\textbf{Subtask} & \textbf{Failure Mode} & \textbf{Frequency (\%)} \\
\midrule
\multirow{3}{*}{\textbf{Moving to Intersection}}
 & Misestimate the distance to the intersection & 53.33 \\
 & Fail to detect the intersection              & 28.33 \\
 & Misidentify the reference landmark           & 18.33 \\
\midrule
\multirow{3}{*}{\textbf{Turning}}
 & Misinterpret the turning pattern             & 42.86 \\
 & Misunderstand history status summary         & 42.86 \\
 & Fail to detect upfront buildings             & 14.29 \\
\midrule
\multirow{3}{*}{\textbf{Reaching Destination}}
 & Fail to match the landmark in a different perspective & 60.00 \\
 & Stop too early to face the landmark          & 30.00 \\
 & Fail to align the landmark                   & 10.00 \\
\bottomrule
\end{tabular*}
\end{threeparttable}
\vspace{-10pt}
\end{table}

\textbf{Baselines.}
 Inspired by the recent success of large vision-language models (VLMs) on navigation tasks~\cite{wang2024qwen2vlenhancingvisionlanguagemodels}, we evaluate multiple recent VLMs as backbones using ReAct~\cite{yao2023reactsynergizingreasoningacting}, including GPT-4o, GPT-o3, GPT-o3-pro \cite{4o}, Gemini 2.5 Flash~\cite{gemini}, Qwen-VL 2.5~\cite{bai2025qwen25vltechnicalreport}, Gemma 3~\cite{gemmateam2025gemma3technicalreport} and InternVL \cite{internvl2.5}. Additionally, we finetune QwenVL2.5-7B on SimWorld-20k. We also test a hybrid baseline, HybridGPT, where GPT-4o is used as a high-level decision maker and A*~\cite{Hart1968} is used as a low-level motion planner. For RL baselines, we train a multimodal policy model, VLA-RL, with DeBERTa-v3~\cite{he2023debertav3improvingdebertausing} as language encoding and DINOv2~\cite{oquab2024dinov2learningrobustvisual} as visual encoding, following VLN-CE~\citep{krantz2020navgraphvisionandlanguagenavigationcontinuous}. We include more baseline implementation details in Appendix~\ref{sec:app_MMNAv_baselines}.




%

\subsection{Results}

\textbf{Zero-shot VLMs.} As shown in Table~\ref{tab:mmnav}, among zero-shot ReAct models, Gemini 2.5 Flash achieves the highest progress score. GPT-4o exhibits a mismatch between its distance progress and subtask completion rate, primarily due to its inability to detect termination conditions, often overshooting the goal and yielding zero distance progress. QwenVL2.5-72B ranks highest among open-source models, while QwenVL2.5-7B performs comparably to significantly larger models. 

\textbf{Finetuned Models.} After fine-tuning, QwenVL2.5-7B shows substantial improvements across all metrics and is the only model to achieve a non-zero full task success rate. However, the absolute success rate remains low, partly because the training set is grounded on oracle action traces, which limits robustness. Incorporating reinforcement learning or corrective demonstrations could further enhance performance.

\textbf{Reasoning Models.} All zero-shot non-reasoning models score zero in SR, highlighting a fundamental capability gap. Case studies in Appendix \ref{sec:app_QE_MMNav} suggest that these models often fail in task completion due to insufficient instruction grounding or inability to handle long-horizon dependencies. However, the results for reasoning models indicate that improved reasoning abilities boost performance. In our experiment, the reasoning models show improved depth estimation and destination alignment, which further demonstrates the importance of visual and spatial reasoning in our benchmark.

\textbf{Other Baselines.} The hybridGPT handles local turning more stably than zero-shot GPT-4o, but lacks fine-grained control, making overall performance more sensitive to GPT-4o's first-attempt accuracy. The RL baseline, VLA-RL, fails to outperform zero-shot LLMs, indicating the difficulty of our benchmark, where sparse reward signals and visually complex spatial reasoning pose challenges for conventional vision encoders. 

\textbf{Hard Setting.} 
We further evaluated realistic obstacle avoidance and traffic rule obedience on models that performed relatively well on the easy setting. As Table \ref{tab:single-agent-hard} shows, 
Gemini 2.5 Flash performs better on avoiding pedestrians and vehicles; however, its red light violation count is higher, not due to failure to stop at red lights, but because the agent often freezes after detecting a red signal, even when already within the intersection. 
This indicates that there is still room for improvement in pragmatic reasoning and safety alignment under real-world conditions. 
Reasoning models, due to their inference latency, are not suitable for this setting of real-time traffic avoidance.

\textbf{Ablation Test.} We construct an ablation test using GPT-4o as the backbone, and we find that the explicit ReAct framework and segmentation provide the most significant marginal improvements among all the components. Details can be found in Table~\ref{tab:ablation} in Appendix~\ref{sec:app_MMNav_quant}.

\textbf{Failure Analysis.} We summarize typical failure modes of VLMs as follows, with specific qualitative examples detailed in Appendix~\ref{sec:app_QE_MMNav}. (1) \textbf{Visual Grounding.} The grounding of VLMs hinder their performance. They fail to recognize the intersection, which is vital in our setting. The relatively low perspective of the robot dog add to the difficulties. (2) \textbf{Spatial and Embodied Reasoning.} VLMs do not yet have good 3D spatial reasoning capacity. They cannot robustly estiamte how close the robot is to a certain intersection or an obstacle. (3) \textbf{Pragmatic Thinking.} To emulate real-world textual navigation, our instructions use high-level phrases like "turn right at the intersection" without detailing the specific turning and crossing actions required. Most models fail to interpret such ambiguity correctly. (4) \textbf{Memory and Planning.} VLMs sometimes fail to adapt the information in memory to the current situation. The frequency for each common failure mode is shown in Table \ref{tab:failure_modes}.

\section{Multi-Robot Search Benchmark}

In many real-world urban applications like search and rescue, multiple robots must collaborate. In large, unfamiliar environments, this requires localizing one another and meeting at a convenient location. Such multi-robot search is foundational for effective collaboration, but presents unique challenges: (i) each robot has partial, egocentric perception, (ii) the environment is dynamic and safety-critical, and (iii) coordination depends on grounded natural-language dialogue. As shown in Table~\ref{tab:multiagent_benchmark_comparison} (Appendix~\ref{sec:compare_mrs}), prior benchmarks only address parts of this problem. We introduce \textsc{SimWorld-MRS}, a benchmark to fill this gap.



\subsection{Setup}

\textbf{Task Definition.} There are two robots, a main robot and a follower robot. The main robot has explored the city, so it has the memory of a map and images of landmarks (typically over 20 landmarks) in the city. However, the follower robot is new to the city and does not have such information. Neither robot knows the other's location. Their goal is to meet each other as soon as possible by physical navigation as well as verbal communication. A task is considered successful when at least one of the robots confirms that it can see the other robot in its egocentric view.

Both robots have similar observation and action spaces as in the SimWorld-MMNav benchmark. Additionally, both robots can send natural language messages to each other. Each robot can send a confirmation signal whenever it believes that it has seen the other robot. 

We provide more details of the benchmark in Appendix~\ref{sec:app_MRS}.



\textbf{Statistics.}
For evaluation in the multi-robot search benchmark, we construct 100 unique urban environments, each covering an area of $2.5\,\mathrm{km}^2$. In each environment, 20 distinguishable landmarks are selected—distributed across all the city blocks—as the main robot's memory of the city. On average, the initial distance between the two robots’ spawning locations is $576\,\mathrm{m}$, requiring 287 steps for an oracle planner to complete the task.






\textbf{Evaluation Metrics.}
We assess multi-agent navigation performance through two principal metrics: (1) \textbf{Collaborative Success Rate (CSR)} \cite{lowe2017multi} measures the percentage of tasks completed successfully before meeting; (2) \textbf{Task Progress} averages, across robots, the fraction of their shortest-path distance that is covered by the end of an episode. 

\textbf{Baselines.} Following the collaboration paradigm of \textsc{RoCo} \cite{mandi2023rocodialecticmultirobotcollaboration}, we enable robots to discuss a joint plan for concrete rendezvous behavior using a VLM. The follower robot will first describe its location using language for the main robot to localize it. Afterwards, two robots will confirm a plan, where the main robot will describe paths to a meeting location for the follower robot to follow, while the main robot will plan its own path to reach the meeting location. In \textsc{SimWorld-MRS}, we first evaluated GPT-4o and Gemini 2.5 Flash with oracal planner and then picked the best performing VLM (GPT-4o) and paired it with RoCo. We include more implementation details in the Appendix~\ref{sec:app_MRS_Baselines}.

\subsection{Results}

\begin{table}[t!]
    \centering
    \caption{Experiment results on \textsc{SimWorld-MRS} benchmark.}
    \label{tab:multi-benchmarks}
    \scriptsize
    \setlength{\tabcolsep}{16pt}
    \resizebox{\textwidth}{!}{%
    \begin{tabular}{llccccc}
    \toprule
    \textbf{Models} & \textbf{Method} & CSR\%$\uparrow$ & Task Progress\%$\uparrow$  \\
    \midrule
    \rowcolor{gray!10} \multicolumn{6}{l}{\hspace{1mm}\textit{Proprietary Models}} \\
     GPT-4o & Oracle Planner & 65.00 & 76.90   \\
     Gemini 2.5 Flash & Oracle Planner & 54.55 & 75.84 &    \\
     GPT-4o & RoCo & 33.33 & 22.93   \\
    \rowcolor{gray!10} \multicolumn{6}{l}{\hspace{1mm}\textit{Open-sourced Models}} \\
    QwenVL 2.5 72B & RoCo & 11.11 & 35.94 & \\
    \bottomrule
    \end{tabular}
    }
\vspace*{-10pt}
\end{table}



Table~\ref{tab:multi-benchmarks} summarizes baseline results on our multi-robot search task, with specific qualitative examples detailed in Appendix~\ref{sec:app_QE_MRS}. GPT-4o with the oracle planner achieves the highest CSR (52\%) and task progress (68.44\%) by combining precise landmark localization with optimal A* planning, showing upper-bound performance with full map access. The RoCo policy lets the follower describe its view, which the map-aware robot localizes via VLM-based retrieval. This one-shot communication enables concurrent movement and more realistic coordination. However, converting rendezvous plans into language introduces ambiguity and execution noise, often causing path deviations. Without iterative replanning, grounding or control errors can't be corrected. Consequently, GPT-4o under RoCo achieves only 33.33\% CSR and 22.93\% task progress—significantly lower than the oracle baseline.


\section{Conclusion}

We have created a novel embodied AI simulator, SimWorld-Robotics~(\simulationname{}), for synthesizing photorealistic and dynamic urban environments. It can procedurally generate infinite photorealistic urban environments. Additionally, it can populate the environments with pedestrians, vehicles, and robots. By leveraging these features, we have built two new robot benchmarks. One focuses on multimodal robot navigation (\textsc{SimWorld-MMNav}), and the other evaluates multi-robot collaboration in searching tasks (\textsc{SimWorld-MRS}). Our experimental results reveal significant limits in strong VLM-based baselines. Our evaluation also demonstrated the value of finetuning VLMs on large-scale training sets synthesized in our simulator.

\textbf{Limitations and Future Work.} Our current simulator focuses only on outdoor environments. The action space of the human agents, though more diverse than prior simulators, is still limited. In the future, we intend to scale up the action space by leveraging recent human body motion generation models. We also plan to incorporate indoor scenes into \simulationname{}.

\bibliographystyle{plain}
\bibliography{reference}

\clearpage
\appendix
\section{SimWorld-Robotics Details}\label{sec:app_simulator}
\vspace{-5pt}
\subsection{Assets}
\begin{figure}[H]
    \centering
    \includegraphics[angle=270, trim=150 135 165 70, clip, width=\linewidth]{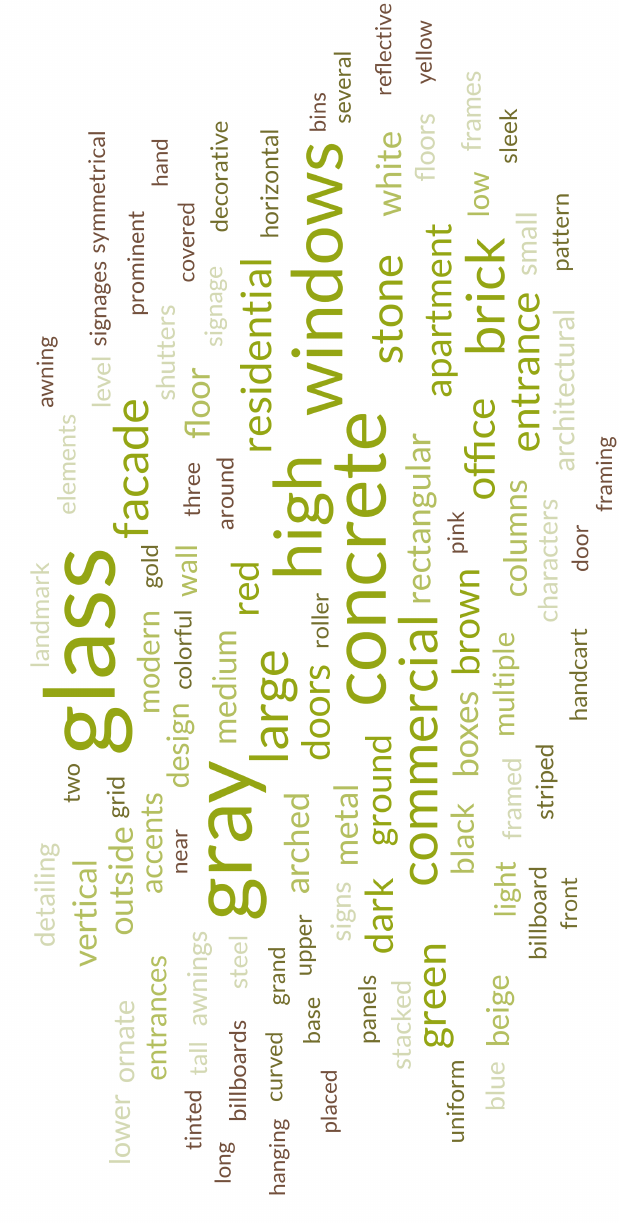}
    \caption{
    \textbf{Building Attribution in \simulationname{}} Each building asset in \simulationname{} is paired with a ground-truth description initially generated by language model and later verified and corrected by human in our team.
    \label{fig:building-word claud}}
    \vspace{-10pt}
\end{figure}

\vspace{-5pt}
\paragraph{Building and Detail Assets}
We utilize a wide range of high-fidelity buildings and street-level details, all sourced from Unreal Engine’s official marketplace, Fab.com. All assets are used in compliance with their respective licenses and terms of use. The ground-truth attribution for building assets are made into word cloud, shown in Figure~\ref{fig:building-word claud}.
\vspace{-10pt}
\vspace{-10pt}
\paragraph{Embodied Agent Assets}
Quadruped robot models are sourced from CGTrader and integrated into \simulationname{} in compliance with the platform's licensing terms. Vehicles are adapted from the official Unreal Engine package \texttt{CitySampleVehicle}, with customized blueprint logic to support the specific requirements of our simulation. Human agents are created using Unreal Engine’s MetaHuman framework, providing diverse, realistic character models. 
\vspace{-5pt}
\subsection{Full Comparison of \simulationname{} with Prior Simulators}\label{sec:app_compare_simulator}
\vspace{-10pt}
\begin{table}[H]
    \begin{threeparttable}
    \centering
    \scriptsize
    \setlength{\tabcolsep}{5.2pt} 
    \renewcommand{\arraystretch}{1.2} 
    \vspace{-5pt}
    \caption{Comparison of simulation platforms across key features. The \textbf{Scenes} section includes environment \text{Type} (U: Urban, I: Indoor), support for \text{Procedural Generation} ($\checkmark$: supported, $\times$: not supported), and level of \text{Photorealism}. The \textbf{Human} section summarizes scene interaction capabilities (\text{Scene-Interact.}) and the number of supported human \text{Actions}. \textbf{Embodied Agents} indicate whether robots (Rob.), humans (Hum.), and vehicles (Veh.) are supported. \textbf{AMC} denotes support for asynchronous multi-agent control.}
    \begin{tabularx}{\textwidth}{lcccccccccc@{}}
        \toprule
        \multirow{2}{*}{Simulator} & \multicolumn{3}{c}{Scenes} & \multicolumn{2}{c}{Human} & \multicolumn{3}{c}{Embodied Agents} & \multirow{2}{*}{AMC} \\ 
        \cmidrule(lr){2-4} \cmidrule(lr){5-6} \cmidrule(lr){7-9} 
        & Type & Procedural Generation & Photorealistic & Scene-Interact. & Actions & Rob. & Hum. & Veh. \\ 
        \midrule
        Matterport3d~\cite{chang2017matterport3dlearningrgbddata}& I & $\times$ & $\checkmark$ & $\times$  & $\times$ & $\times$  & $\times$  & $\times$  &  $\times$  \\
        Sean2.0~\cite{tsoi2022sean2}        & I & $\times$ & $\times$ & $\times$  & 3 & $\checkmark$  & $\times$  & $\times$  &  $\times$  \\ 
        Arena3.0~\cite{kästner2024arena30advancingsocial}       & I & $\checkmark$ & $\times$ & $\times$  & 8 & $\checkmark$  & $\times$  & $\times$  &  $\times$  \\ 
        AI2THOR~\cite{kolve2022ai2thorinteractive3denvironment}        & I & $\checkmark$ & $\times$ & $\times$  & $\times$ & $\checkmark$  & $\times$  & $\times$  & $\checkmark$  \\ 
        TDW~\cite{gan2021threedworldplatforminteractivemultimodal}    & I & $\checkmark$ & $\checkmark$ & $\times$  & 50 & $\checkmark$  & $\times$  & $\times$  &   $\times$  \\      
        SocNavBench~\cite{biswas2021socnavbenchgroundedsimulationtesting}    & I & $\times$ & $\times$ & $\times$ & 2 & $\checkmark$  & $\times$  & $\times$  &  $\times$  \\   
        Habitat 3.0~\cite{puig2023habitat30cohabitathumans}     & I & $\checkmark$ & $\times$ & $\checkmark$  & 4 & $\checkmark$  & $\checkmark$  & $\times$  &  $\checkmark$ \\   
        VirtualHome 2.0~\cite{puig2021watchandhelpchallengesocialperception}    & I & $\checkmark$ & $\times$ & $\checkmark$  & 25 & $\times$  & $\checkmark$  & $\times$  & $\times$  \\   
        BEHAVIOR~\cite{pmlr-v205-li23a}  & I & $\checkmark$ & $\checkmark$ & $\times$  & $\times$ & $\checkmark$  & $\times$  & $\times$  &  $\times$  \\  
        CARLA~\cite{Dosovitskiy17} & U & $\times$ & $\checkmark$ & $\times$  & 2 & $\times$  & $\checkmark$  &  $\checkmark$  &  $\checkmark$  \\ 
        AirSim~\cite{shah2017airsimhighfidelityvisualphysical}& U & $\times$ & $\checkmark$ & $\times$  & 2 & $\times$  & $\checkmark$  &  $\checkmark$  &  $\checkmark$ \\ 
        MetaUrban~\cite{wu2024metaurbanembodiedaisimulation}      & U & $\checkmark$ & $\times$ & $\times$  & 2 & $\checkmark$  & $\times$  & $\times$  & $\checkmark$  \\ 
        EmbodiedCity~\cite{gao2024embodiedcitybenchmarkplatformembodied}   & U & $\times$ & $\times$ & $\times$  & 2 & $\checkmark$  & $\times$  & $\times$  &  $\times$  \\ 
        Grutopia~\cite{wang2024grutopiadreamgeneralrobots}       & U & $\times$ & $\checkmark$ & $\times$  & $\times$ & $\checkmark$  & $\times$  & $\times$  & $\times$  \\ 
        AerialVLN~\cite{liu2023aerialvlnvisionandlanguagenavigationuavs}      & U & $\times$ & $\checkmark$ & $\times$  & $\times$ & $\checkmark$  & $\times$  & $\times$ & $\times$  \\   
        \textbf{\simulationname{} (Ours)}  & U & $\checkmark$ & $\checkmark$ & $\checkmark$  & 26 & $\checkmark$  & $\checkmark$  & $\checkmark$  & $\checkmark$  \\ 
        \bottomrule
    \end{tabularx}%
    \label{tab:simulators_comparison_full} 
    \end{threeparttable}
    \vspace{-10pt}
\end{table}

\subsection{Asynchronous Multi-agent Control}\label{sec:app_AMC}

\begin{figure}[H]
    \centering
    \includegraphics[trim=0 170 0 140, clip,width=\linewidth]{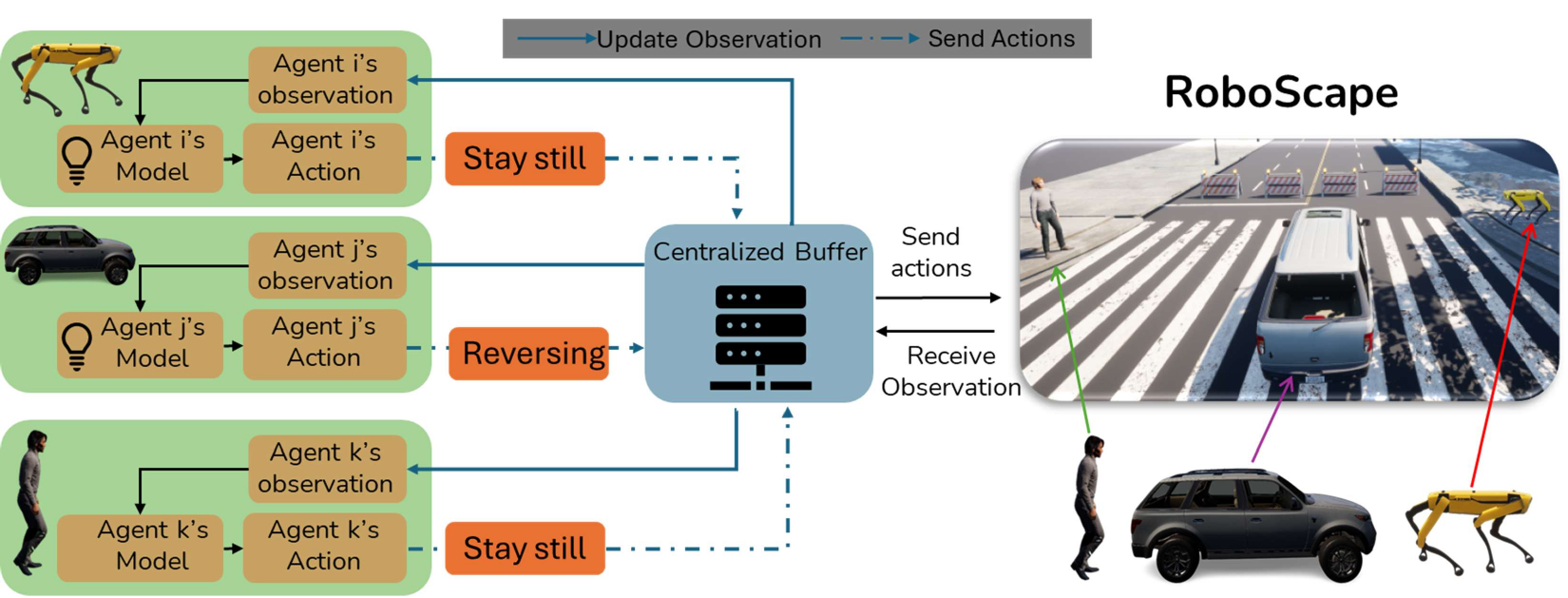} 
    \caption{\textbf{Asynchronous multi-agent control.} To ensure each agent independently selects and executes actions based on its local observation and policy model, a centralized buffer serves as the interface between agents and the environment, receiving actions from all agents and updating their observations based on the resulting environmental changes. } 
    \vspace*{-5pt}
    \label{fig:aynchronized multi-agent control} 
\end{figure}

The control pipeline is illustrated in Figure~\ref{fig:aynchronized multi-agent control}, where each agent—regardless of embodiment (robot, vehicle, or pedestrian)—independently perceives its local observation and selects an action using its policy model. These actions are asynchronously sent to a centralized buffer, which mediates the interaction between agents and the environment by updating each agent’s observation based on the environment’s response to all executed actions.

\subsection{Observation Space}\label{sec:app_obs}
\vspace*{-5pt}

\begin{figure}[H]
    \centering
    \includegraphics[trim=2.8cm 5cm 2.4cm 4.5cm, clip, width=\linewidth]{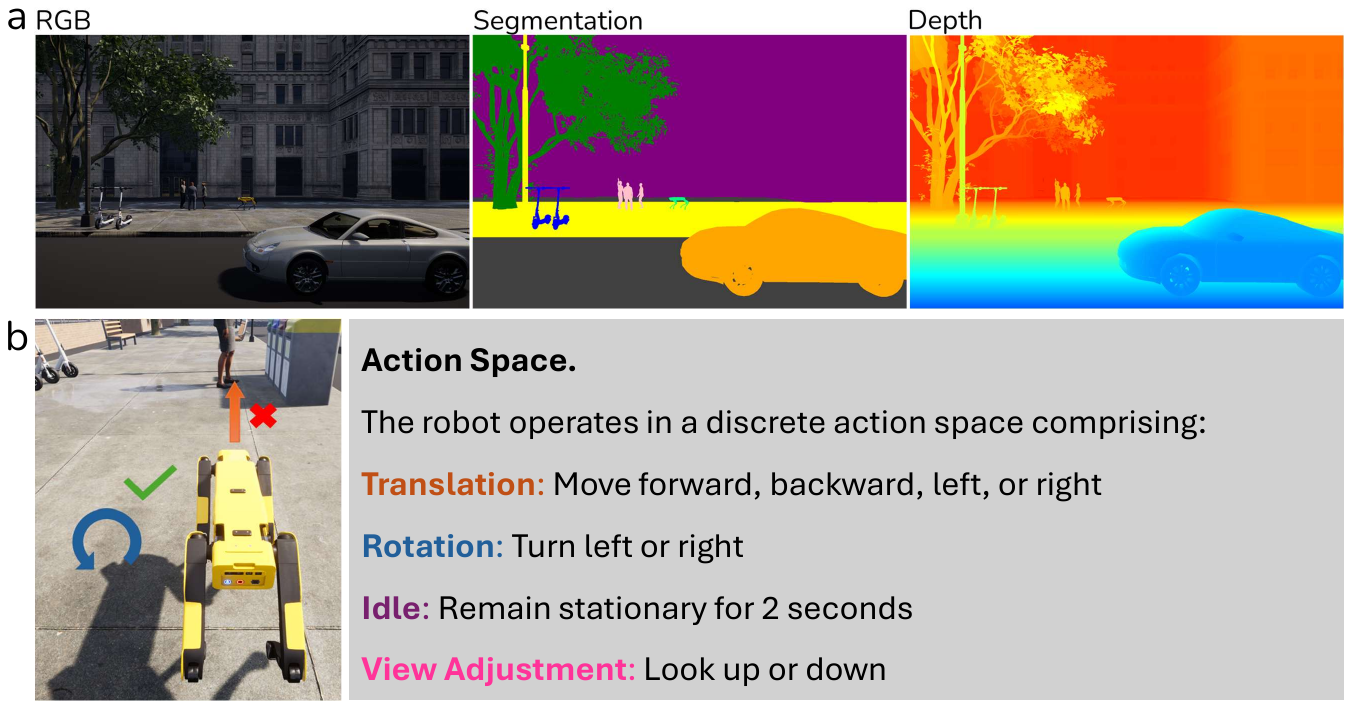} 
    \caption{\textbf{Multimodal observation and discrete action space}.(a) The robot perceives the environment via RGB, segmentation, and depth modalities. (b) Its action space includes translation, rotation, idling, and view adjustment.} 
    \label{fig:observation-and-action} 
    \label{fig:observation space}
\end{figure}

As illustrated in Figure~\ref{fig:observation-and-action}a, the robot receives multimodal observations at each step, including RGB images, semantic segmentation maps, and depth maps, enabling a rich understanding of its surrounding environment.
\vspace*{-10pt}
\vspace*{-5pt}
\subsection{Robot Action Space}\label{sec:app_action_space}
\vspace*{-5pt}

The robot executes actions within a discrete control space composed of directional movements, rotations, idle behavior, and vertical view adjustments, as illustrated in Figure~\ref{fig:observation-and-action}b.

\subsection{Procedural City Generation}\label{sec:app_city_generation}
The Procedural City Generation pipeline is a central component of the \simulationname{} simulator, designed to generate realistic and structured urban environments from minimal input specifications. This pipeline offers a highly modular and extensible architecture that supports the creation of diverse cityscapes, enabling a wide range of embodied AI tasks such as autonomous driving, pedestrian navigation, and multi-agent simulations.

As illustrated in Figure~\ref{fig:procedural generation}, the city generation process is organized into four sequential stages: road generation, building generation, street element generation, and traffic element generation. Each stage progressively adds layers of realism and complexity to the simulated environment.

\textbf{Road Generation}: The process begins with the creation of a road network, which serves as the backbone of the city. Roads are generated through an initiation phase and a tree-like growth process that balances depth and branching using a priority queue. Mechanisms such as road-end attachment and intersection checking ensure a coherent and plausible layout.

\textbf{Building Generation}: Once roads are established, buildings are procedurally placed along road segments. For each side of the road, candidate positions are sampled while checking for space availability and avoiding collisions. A greedy strategy is used to fill remaining gaps near road ends, maximizing spatial utilization and maintaining visual uniformity.

\textbf{Street Element Generation}: Smaller environmental elements such as trees, road cones, benches, and parked vehicles are generated around buildings and alongside roads. These details are categorized and placed based on contextual zones—either surrounding buildings or within designated sidewalk areas. While collisions with other objects are not strictly enforced for performance reasons, the placement respects basic accessibility constraints.

\textbf{Traffic Element Generation}: The final stage involves populating the city with dynamic actors such as cars, pedestrians, and agents. These elements bring life to the simulation and interact with the static environment, enabling research in traffic flow, behavioral modeling, and agent-based navigation.

Internally, the pipeline utilizes dedicated managers for roads, buildings, and elements. Each manager maintains spatial data using both lists and quadtree structures—enabling efficient queries, spatial indexing, and collision checks. Procedural rules and constraints guide item generation to ensure the city is functionally consistent and visually appealing.

By separating generation into clearly defined stages and maintaining a rule-driven architecture, this pipeline provides a robust foundation for scalable and customizable city simulation within \simulationname{}.

\subsection{Background Traffic System}\label{sec:app_traffic_system}
\begin{figure}[H]
    \centering
    \includegraphics[trim=3cm 8cm 3cm 7cm, clip, width=\linewidth]{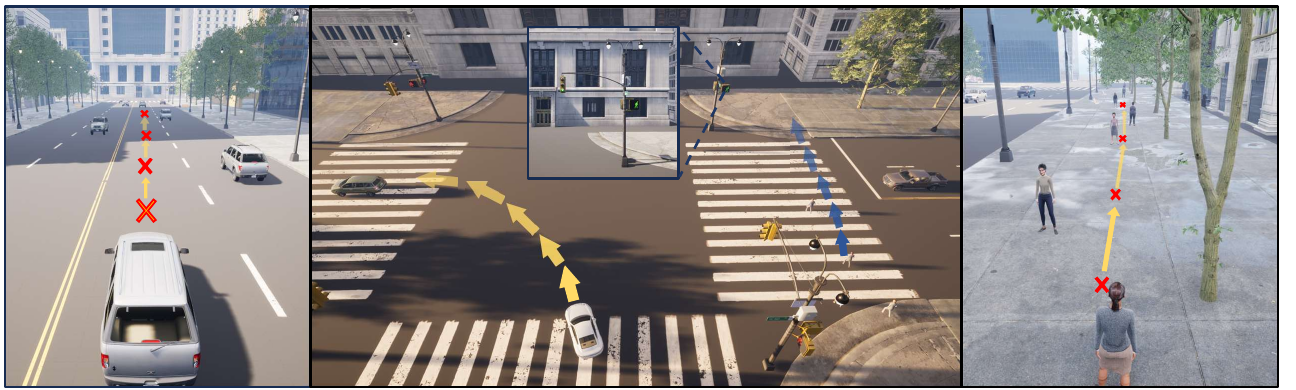} 
    \caption{\textbf{Traffic system in \simulationname{}.} Agents follow sampled waypoints and obey traffic lights at intersections, proceeding only if it's green signal.} 
    \label{fig:traffic system} 
\end{figure}

\vspace*{-5pt}
In \simulationname{}, both pedestrians and vehicles are controlled using a rule-based traffic system, as illustrated in Figure~\ref{fig:traffic system}. A waypoint system is constructed over the entire procedurally generated city, consisting of two types of waypoints: road waypoints and intersection waypoints. At each intersection, four intersection waypoints are sampled, one at each corner. For every road segment connecting two intersections, road waypoints are sampled at 17-meter intervals, linking the intersection waypoints at both ends. Prior to traffic simulation, we sample a sequence of connected waypoints for each pedestrian and vehicle. During simulation, each agent follows its assigned path. When an agent reaches an intersection waypoint, it checks the traffic light status: if the light is green and the remaining duration exceeds 15 seconds, the agent proceeds; otherwise, it waits until the signal turns green, ensuring safe and realistic traffic behavior.

\subsection{Customization in \simulationname{}}\label{sec:app_customization}
The two benchmark tasks included in the paper are intended as case studies to demonstrate the functionalities and utilities of \simulationname{}. \simulationname{} was explicitly designed to support user-friendly customization of embodied agents, environments and tasks.
\subsubsection{New Embodied Agents}
To add a new robot type in \simulationname{}, one can leverage the existing Python API of \simulationname{} to conveniently customize action spaces—either continuous or discrete—as well as observation spaces such as RGB, depth, or semantic segmentation images. The required additional work (1) obtaining a new robot asset, typically from the Unreal Engine Marketplace; (2) defining the robot's actions using Unreal's Blueprint system; (3) integrating these actions with our Python API to enable high-level control; and (4) attaching our camera components to support the desired observation space.

\subsubsection{New Environments}
Users can generate diverse and realistic urban layouts through our Python API by providing simple metadata inputs (e.g., number of streets, street length, object categories, and their spatial distributions, such as “10\% trees, 5\% tables and chairs”). This allows researchers to create arbitrarily large and varied city environments.

\subsubsection{New Tasks}
Defining tasks in \simulationname{} is similar to standard Gym environments~\cite{brockman2016openaigym}, as the Python API of \simulationname{} follows the same format for agent control (including pedestrians and vehicles). Users can spawn different types of agents, customize observation spaces (e.g., RGB, depth, or semantic segmentation) and action spaces (continuous or discrete actions), and program new goal definitions (e.g., language instructions, target images, or spatial objectives). Additionally, while the current pedestrians follow rule-based logic, users can override behaviors to simulate complex or rare cases. For instance, one can simulate a jaywalking pedestrian by scripting a few agents to cross during a red light, while placing a robot at the intersection to evaluate its reaction.

\subsection{\simulationname{} Technical and Operational Details}

\textbf{Underlying simulation engine.} Our simulation is built on UE5(Unreal Engine 5), leveraging its native Chaos physics pipeline: each object is assigned an appropriate collision mesh, and at each fixed simulation tick performs discrete‑time integration of Newton’s equations—resolving forces, collisions, and joint constraints via its iterative solver.

\textbf{Python API and runtime environment.} On top of UE5, we’ve implemented a dedicated Python layer that communicates with the engine through an updated UnrealCV‑based TCP server~\cite{10.1145/3123266.3129396}, where high‑level commands issued in Python are forwarded over TCP straight into the UE5 runtime. On top of this API, we provide a standard Gym interface so that researchers can plug in and benchmark any baseline with minimal effort. We will distribute a Windows executable and a Singularity container for Ubuntu and macOS so that users can run \simulationname{} out of the box without local compilation; all binaries, container images, and the Gym wrapper will be open‑sourced upon publication.

\textbf{Human-in-the-loop interface.} We support a human-in-the-loop interface through which a human operator uses a mouse and keyboard to control the robot; all trajectories are recorded automatically as expert demonstrations for downstream training.

\subsection{\simulationname{} Computational Requirements}
To ensure wide accessibility, \simulationname{} supports adjustable rendering resolutions, allowing deployment across both high-end servers and modest laptops. 

\textbf{Recommended Setup} 
\vspace{-5pt}
\begin{itemize}
    \item \textbf{CPU:} Intel Core i7-12700H or AMD Ryzen 9 5900HS
    \item \textbf{GPU:} NVIDIA RTX 3070 or GPU with more than 6 GB
    \item \textbf{RAM:} 32 GB
\end{itemize}
\vspace{-5pt} 
\textbf{Minimum Setup (60 FPS for \simulationname{})}
\vspace{-5pt} 
\begin{itemize}
    \item \textbf{CPU:} Intel Core i7-11300H or AMD Ryzen 9 4800H
    \item \textbf{GPU:} NVIDIA RTX 2060 (notebook)
    \item \textbf{RAM:} 16 GB
\end{itemize}

\subsection{Running Efficiency of \simulationname{}}
We evaluated all baselines on a headless machine with an AMD EPYC 9534 CPU, L40S GPU and 64 GB RAM. We can run 2 instances in parallel with a fixed 60 fps. Here's the table when we stress-test the runtime performance with different settings.
\vspace{-15pt} 
\begin{table}[H] 
    \centering
    \caption{Runtime Performance Stress Test} 
    \label{tab:performance}
    \centering
    \scriptsize
    \setlength{\tabcolsep}{16pt}
    \resizebox{\textwidth}{!}{%
    \begin{tabular}{lllll} 
        \toprule 
        \textbf{Resolution} & \textbf{Rendering Quality} & \textbf{GPU Utils (\%)} & \textbf{CPU Utils (\%)} & \textbf{RAM (MB)} \\
        \midrule 
        $640 \times 360$ & low & 30.04 & 16.41 & 561.56 \\
        $720 \times 600$ & low & 29.72 & 17.73 & 596.4 \\
        $1280 \times 720$ & low & 26.5 & 18.56 & 734.81 \\
        \midrule 
        $640 \times 360$ & high & 30.03 & 16.59 & 564.04 \\
        $720 \times 600$ & high & 29.61 & 17.47 & 589.3 \\
        $1280 \times 720$ & high & 27.44 & 19.08 & 738.6 \\
        \bottomrule 
    \end{tabular}
    }
\end{table}
\section{Benchmark Comparison}

\subsection{Comparing SimWorld-MMNav with Prior Vision-Language Navigation Benchmarks}\label{sec:compare_nav}

\begin{table}[H]
 \caption{Comparison of instruction following benchmarks for navigation.}
 \scriptsize
\setlength{\tabcolsep}{2pt} 
    \centering
    \resizebox{\textwidth}{!}{%
    \begin{tabular}{lccccccc}
        \toprule
        Benchmark & \multicolumn{2}{c}{Env} & \multicolumn{2}{c}{Agent} & \multicolumn{3}{c}{Route} \\
        \cmidrule(lr){2-3} \cmidrule(lr){4-5} \cmidrule(lr){6-8} 
        & Type & Num & Type & Act Space & Gen & Acts & Num  \\
        \midrule
        CVDN~\cite{thomason2019visionanddialognavigation} &Indoor, Static & 90 & Camera & Graph-based & Manual & 7 & 7,415 \\
        REVERIE\cite{qi2020reverieremoteembodiedvisual} &Indoor, Static & 90 & Camera & Graph-based & Manual & 5 & 7,000  \\
        TouchDown \cite{chen2020touchdownnaturallanguagenavigation} & Outdoor, Static & 1 & Camera & Graph-based & Procedural & 35 & 9,326  \\
        ANDH \cite{fan2023aerialvisionanddialognavigation}& Outdoor, Static & 1 & Drone & 6 & Manual & 7 & 6,269\\
        AerialVLN \cite{liu2023aerialvlnvisionandlanguagenavigationuavs}& Outdoor, Static & 25 & Drone & 8 & Manual & 204 & 8,446 \\
        \textbf{SimWorld-MMNav (Ours)} & Outdoor, Dynamic & 400 & Robot & 7 & Procedural & 100 & 400 \\
        \bottomrule
    \end{tabular}
    }
   
    \label{tab:benchmark_comparison}
\end{table}

\subsection{Comparing SimWorld-MRS with Prior Multi-Robot Collaboration and Navigation Benchmarks}\label{sec:compare_mrs}

\begin{table}[h]
\caption{Comparison of multi-robot collaboration and navigation benchmarks.  
“Dynamic” indicates whether pedestrians, vehicles, or other moving obstacles are present.}
\setlength{\tabcolsep}{2pt}  
\centering
\resizebox{\textwidth}{!}{%
\begin{tabular}{lccccccccc}
\toprule
\textbf{Benchmark} 
& \multicolumn{2}{c}{\textbf{Env}} 
& \multicolumn{2}{c}{\textbf{Agent}} 
& \multicolumn{2}{c}{\textbf{Comm.}} 
& \textbf{Map} 
& \textbf{Objective} \\

\cmidrule(lr){2-3}\cmidrule(lr){4-5}\cmidrule(lr){6-7}
& Type & Scale 
& Num & Type 
& Modality & Max Len 
& Access & \\

\midrule
RoCo \cite{mandi2023rocodialecticmultirobotcollaboration}           
& Indoor, Static & Room           
& 2 & Mobile arm      
& None & --            
& None & Collab.\ pick\&place \\

DOROTHIE \cite{ma2022dorothiespokendialoguehandling}     
& Outdoor, Dynamic & Road           
& 1 (ego car) & Vehicle        
& Spoken dlg. & $\sim\!20$ tok.  
& Full & Dialogue nav. \\

MetaUrban \cite{wu2024metaurbanembodiedaisimulation}  
& Outdoor, Dynamic & City           
& 1\,+ & E-scooter/robot     
& None & --            
& Full & Micromobility nav. \\

DriVLMe \cite{huang2024drivlmeenhancingllmbasedautonomous}         
& Outdoor, Dynamic & City           
& $10+$ & Vehicle         
& None & --            
& Full & Coop.\ driving \\

RobotSlang \cite{banerjee2020robotslangbenchmarkdialogguidedrobot} 
& Indoor, Static & Lab          
& 2–3 & Mobile base    
& Natural-lang. & 80\,char
& Partial & Joint object search \\

Where Are You? \cite{hahn2021youlocalizationembodieddialog}   &
Outdoor, Static & City &
2 (tourist+guide) & Pedestrian &
Natural-lang. & 40\,tok. &
Split & Localization via dialog \\

\textbf{SimWorld-MRS (Ours)}          
& Outdoor, Dynamic & City-scale          
& 2 & Robot            
& Natural-language & 128 char 
& Split & Rendezvous / meet-up \\

\bottomrule
\end{tabular}}

\label{tab:multiagent_benchmark_comparison}
\end{table}
\section{\textsc{SIMWORLD-MMNav}}\label{sec:app_MMNAv}

In this section, we present \textbf{SimWorld-MMNav}, a single-robot benchmark designed to evaluate multimodal navigation in large-scale urban environments. We begin by describing our procedural task generation pipeline, which enables the creation of diverse and realistic navigation tasks. We then introduce the \textbf{SWR-20k} training dataset, detailing how we generate fine-grained supervision signals for multimodal learning. Next, we elaborate on the baseline models used in our experiments, including their implementation pipelines and prompting strategies. Finally, we provide an in-depth analysis of notable failure cases observed during evaluation, such as incorrect 3D spatial reasoning and other representative errors.

\subsection{Detailed Task Settings}

\paragraph{Observation Space} In addition to the multimodal instruction, the robot is equipped with a compass that indicates its current facing direction. To facilitate navigation, the robot also receives its egocentric RGB image, segmentation image, and depth image, which together provide a rich perception of the surrounding environment.

\paragraph{Action Space} The robot’s action space consists of two categories: \textbf{movement actions} and \textbf{task-related actions}. In the movement category, the robot can choose from the following options: move forward, move backward, move left, move right, turn left ($-90^\circ$), turn right ($+90^\circ$), or stay still. In the task-related category, the robot can select the action \texttt{evaluate}, indicating that it believes the current task has been completed. If the evaluation is correct, the robot receives the next subtask’s language instruction and visual hint; otherwise, the episode terminates as a failure. Our simulator allows active perception like looking up by providing egocentric observation from different fields of view.

\subsection{Metric Detail}

We evaluate Single-Agent Instruction Following using three metrics: Success Rate (SR), Subtask Success Rate (SSR), and Distance Progress (DP).

\paragraph{Success Rate} This metric measures the proportion of navigation tasks in which the agent successfully reaches the final destination. A trial is counted as successful only if the agent stops near the correct landmark building and is oriented towards it at the end of the trajectory.

\paragraph{Subtask Success Rate} This metric evaluates the proportion of correctly completed subtasks within a navigation episode. Let $N$ denote the total number of subtasks in a given instruction, and $n_c$ the number of subtasks successfully completed by the agent. The Subtask Success Rate (SSR) for that episode is computed as:
\begin{equation}
\text{SSR} = \frac{n_c}{N}
\label{eq:ssr}
\end{equation}
We report the average SSR across all test episodes.

\paragraph{Distance Progress} This metric quantifies how much closer the agent is to the goal at the end of the navigation compared to the beginning. Let $d_0$ be the initial distance to the goal and $d_T$ the final distance. The Distance Progress (DP) is computed as:
\begin{equation}
\text{DP} = \max\left(\frac{d_0 - d_T}{d_0}, 0\right)
\label{eq:dp}
\end{equation}
DP values are also averaged across all tasks to obtain the final score. All distances are computed using the Manhattan metric.

In the more challenging environment containing obstacles, pedestrians, and vehicles, we additionally measure three metrics to assess the robot's social navigation capabilities: the average number of static collisions (i.e., collisions with buildings or static obstacles), the average number of dynamic collisions (i.e., collisions with pedestrians or vehicles), and the number of actions that violate traffic light rules during navigation.

\subsection{Procedural Task Generation}\label{sec:app_task_generation}

\begin{algorithm}[ht]
\caption{Task Generation for SimWorld-MMNav}
\KwIn{Navigation path $P = \{n_1, n_2, \dots, n_k\}$, orientations $\{\theta_1, \theta_2, \dots, \theta_k\}$}
\KwOut{Ordered list of subtasks $\mathcal{T} = \{(I_1, V_1), (I_2, V_2), \dots\}$}

Initialize subtask list $\mathcal{T} \leftarrow [\,]$\;

$V_1 \leftarrow$ capture visual cue at $n_1$ with orientation $\theta_1$\;

$I_1 \leftarrow$ generate \textbf{Orientation Alignment} instruction using compass heading $\theta_1$ and nearby landmark\;

Add $(I_1, V_1)$ to $\mathcal{T}$\;

\For{$i \leftarrow 2$ \KwTo $k-1$}{
    \If{$\theta_i \neq \theta_{i-1}$}{
        $V_\text{move} \leftarrow$ capture visual cue at $n_{i-1}$ with orientation $\theta_{i-1}$\;
        
        $I_\text{move} \leftarrow$ generate \textbf{Move Along the Road} instruction using a landmark near $n_{i-1}$\;

        Add $(I_\text{move}, V_\text{move})$ to $\mathcal{T}$\;

        $V_\text{turn} \leftarrow$ capture visual cue at $n_i$ with new orientation $\theta_i$\;

        $I_\text{turn} \leftarrow$ generate \textbf{Turn at Intersection} instruction based on the relative angle between $\theta_{i-1}$ and $\theta_i$\;

        Add $(I_\text{turn}, V_\text{turn})$ to $\mathcal{T}$\;
    }
}

$V_\text{goal} \leftarrow$ capture visual cue at $n_k$ with orientation $\theta_k$\;

$I_\text{goal} \leftarrow$ generate \textbf{Reach Destination} instruction using goal landmark description\;

Add $(I_\text{goal}, V_\text{goal})$ to $\mathcal{T}$\;

\Return $\mathcal{T}$
\label{alg:SimWorld-mmnav-task-gen} 
\end{algorithm}

To support diverse multimodal navigation scenarios, we construct the \textsc{SimWorld-MMNav} benchmark under two difficulty levels: \textit{easy} (without obstacles) and \textit{hard} (with obstacles such as vehicles and pedestrians). We generate 200 city maps using our procedural city generation pipeline, with 100 maps allocated to each setting. For the map under hard setting, streetside obstacles are additionally generated. The task generation process is detailed in Algorithm~\ref{alg:SimWorld-mmnav-task-gen}.

On each generated map, we randomly sample a pair of points, denoted as $P_\text{start}$ and $P_\text{goal}$. For each point, we locate the nearest landmark building and extract the front-door location, referred to as $L_\text{start}$ and $L_\text{goal}$ respectively. The robot is spawned at $L_\text{start}$ and tasked with navigating to $L_\text{goal}$.

We then use the A* algorithm to compute an optimal path between $L_\text{start}$ and $L_\text{goal}$ over the city-wide waypoint graph.

This path is decomposed into a sequence of subtasks that reflect the robot’s expected behavior along the route. These subtasks are:

\paragraph{Orientation Alignment}  
Each episode begins with an \textbf{orientation alignment} subtask. The robot is provided with a target compass direction (e.g., North, South, East, West) and a nearby landmark description to assist alignment, such as: “Face north. You will see a modern building with light blue glass on your left.” An image is captured at the initial location with the correct orientation to serve as the visual cue for this step.

\paragraph{Move Along the Road}  
As the robot follows the computed path, we iterate through consecutive waypoints. When the robot is expected to travel straight between two intersections, we define a \textbf{move along the road} subtask. Specifically, we identify a prominent landmark near the intersection where the robot is expected to turn, and use it to generate a descriptive instruction, such as: “Move along the road and stop at the intersection when you see a large building with glass panels and a light brick base on the opposite side.” A visual cue is captured at the intersection with the robot’s current orientation to assist visual grounding.

\paragraph{Turning at Intersections}  
When a change in orientation is detected in the path—typically at an intersection—we insert a \textbf{turn at the intersection} subtask immediately following the previous “move along the road” step. Based on the relative orientation of the next waypoint, we determine whether the robot should turn left or right. A new visual cue is captured at the next node with the updated heading, and a corresponding instruction is generated, such as: “Turn left at the intersection and you should see this view.”

\paragraph{Repeat Navigation Steps}  
The “move along the road” and “turn at the intersection” subtasks may repeat multiple times until the robot reaches the final waypoint along the planned path.

\paragraph{Reach Destination}  
At the final step, we define a \textbf{find destination} subtask. A detailed description of the goal building and its spatial relationship to the robot’s position is used to generate the final language instruction. A visual cue is captured with the robot facing the destination building to aid recognition.

This process completes one full navigation episode. For each map, we generate two episodes by sampling different pairs of start and goal locations, resulting in a total of 400 navigation tasks—200 under the easy setting and 200 under the hard setting.

\subsection{Details of SWR-20k}\label{sec:app_20k}

The SWR-20k is generated as training dataset using a distinct set of maps, referred to as training maps, which differ from those used for evaluation. The primary difference lies in the building distribution: training maps include only 66\% of the building types that appear in the testing maps. The remaining 34\% of building types are held out exclusively for testing, ensuring a clear separation between seen and unseen environments and promoting better generalization.

We generated 100 training maps using only the selected 66\% of building assets. On each map, two tasks are sampled, and the oracle trajectories are generated by A* algorithm, resulting 200 orcale trajectories. Each trajectory contains over 100 steps, forming a training corpus of 20K steps.

At each step during training, the robot receives a synthesized observation and a corresponding sub-instruction, and is required to predict the correct next action. To enhance the robot's ability to reason about the goal and environment, we supervise not only the action prediction but also several intermediate reasoning targets. Specifically, the robot is trained to jointly predict: (1) the distance between the current observation and the target visual hint; (2) the orientation of the visual hint; and (3) the potential sequence of actions from the current location to the final goal. All of these supervisory signals are paired with ground-truth annotations and serve as multi-task learning objectives, guiding the robot toward a deeper understanding of spatial context and task intent.

\subsection{Baseline Detail}\label{sec:app_MMNAv_baselines}

\paragraph{Zero-shot Pipeline} 

\begin{figure}[!h]
    \centering
    \includegraphics[angle=270, trim=80 20 80 30, clip,width=\linewidth]{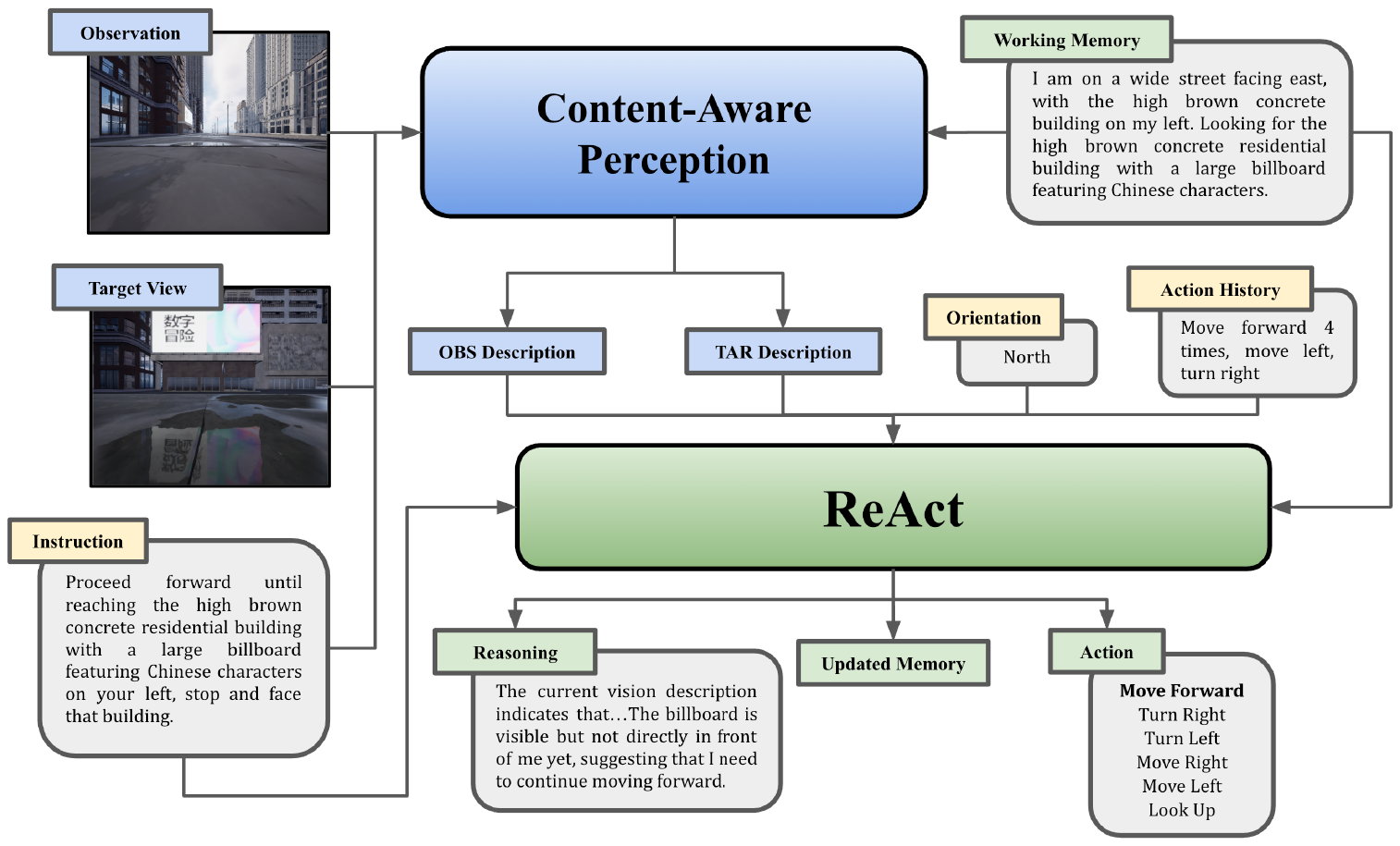}
    \vspace*{-10pt}
    \caption{Zero-shot single agent multimodal instruction following pipeline}
    \label{fig:single-agent baseline}
    \vspace*{-10pt}
\end{figure}

To reduce the mapping complexity and prevent hallucinations, we decompose step-wise action prediction in multimodal instruction-following into two modules: a context-aware perception module and a ReAct-based \cite{yao2023reactsynergizingreasoningacting} action module as detailed in Figure \ref{fig:single-agent baseline}. Throughout this process, the model autonomously maintains a working memory.

The Context-Aware Perception module receives the current observation, a visual hint (i.e., the expected view after subtask completion), the language instruction, and the current memory state. We prompt a vision-language model with the instruction and memory to generate a task-relevant description. This description incorporates information from both the current observation and the target view, and is explicitly encouraged to compare the two images in order to assess whether the subtask has been completed.

The ReAct module operates purely in the text domain. It receives as input the perception-generated description, the original instruction, the memory, the robot’s current orientation, and a textualized history of past actions. ReAct is expected to first perform reasoning, then update the memory module accordingly, and finally select the next action for the robot.

We utilize segmentation images solely as auxiliary visual signals. Given the varying levels of training and understanding of depth images across different models, and considering that interpreting depth information is not the primary focus of our task, we exclude depth images from all inputs to eliminate potential confounding factors. In our experiments, we allow the model to generate a sequence of actions at each step. This design improves execution efficiency during long straight-path phases and serves as a test of the model's short-horizon planning capabilities.

\begin{promptbox}{Content-Aware Perception System Prompt in Zero-Shot Single Agent Instruction Following}
You are a perception module of a navigation robot in a 3D environment. The ultimate goal is to place yourself right in front of a particular building.

You will be given:
- A history summary.
- Your vision description at the last step.
- A single egocentric image.
- The exact expected view you will see once you complete the current subtask.
- A segmentation image. The segmentation of the entire view. Green = trees, purple = buildings, yellow = sidewalks/crosswalks, black = driveways.
- The subtask you are working on.
- The current cardinal direction.

Instructions:
- First, describe the observation in detail, focusing on the color, texture, attachment, etc. of buildings.
- Focus on potential landmarks (if needed), obstacles, or intersections. Reason about useful details for actions.
- Include details like whether the agent is walking straight on a sidewalk and where the driveway is.
- If the upper parts of the entire observation are filled with buildings, the agent is not walking straight along a sidewalk.
- Judge the distance by the vertical position of the object to infer how many forward steps can be taken or how close intersections are.
- Finally describe the expected view, also attentive to the buildings. Match the expected view with current view to determine if the subtask is completed.

When aligning with the last landmarks:
- When the subtask asks you to face a building, first ensure proximity to the landmark, not just visibility in the strip. Look for close-up details to ensure you have reached the right position.
- You don't need to mention the orientation, because it will be given.

Intersection:
- When approaching intersections, actively look for the expected landmark.
- Keep in mind: the landmark might not be visible at the intersection due to limited field of view.
- If you cannot see the landmark, only use the expected view as a reference.
- Turning once alone does not guarantee completion of the turning subtask. Always verify against the expected view.

Output Format:
You must return a JSON object like:
{"Reason": "The detailed description and your reasoning about what detail is useful.",
 "Description": "A useful summary of the observation"}
\end{promptbox}

\begin{promptbox}{ReAct System Prompt in Zero-Shot Single Agent Instruction Following}
You are a navigation robot in a 3D environment. The ultimate goal is to place yourself next to a particular building.

You will be given:
- A list of actions that have already been taken (action history).
- The action sequence you took last step.
- A description of last step's observation. You can use it to compare with the current observation.
- A summary of the history and current status. You can use and update the summary as a hint for future planning.
- The current subtask you are working on.
- Current orientation of the robot.
- A detailed description of the agent's current perception. Then a description of the exact expected view you will see once you complete the current subtask.

Valid actions:
-1: Subtask_completed - If you believe the current subtask is completed, the action sequence should be [-1].
 0: Move_forward - Move 5 meters forward in the direction the robot is facing.
 1: Rotate_left - Rotate 90° to the left.
 2: Rotate_right - Rotate 90° to the right.
 3: Move_left - Move 5 meters left, without rotating.
 4: Move_right - Move 5 meters right, without rotating.

Instructions:
- First, analyze the current visual observation, the instruction, current situation, and history, and reason about how to update the history summary and the next action.
- Next update and resummarize the history and current status. If you have aligned to the current landmark, update the landmark to the next one.
- Finally, decide the next action steps based on the previous analysis.

Alignment:
- You have the cardinal direction to help you align at the beginning of the task.

The last Alignment:
- When handling the "face the building" subtask, you must be close enough and turn to face the building to complete the subtask.
- If you cannot see the building after rotating, it means you are not close enough.

Intersection:
- While reaching intersections, actively look for the expected landmark. Once it's spotted, update your history so the next intersection is the one.
- Keep in mind: the landmark might not be visible at the intersection due to limited field of view—use the expected view as your reference.
- Turning once alone does not guarantee completion of the turning subtask. Always verify against the expected view.

Important Rules:
- Make sure you are oriented along the sidewalk when following "move forward" commands.
- You can plan by outputting variable-length action sequences. For example, [0, 0, 0, 0, 0] if the path is clear.
- Shorter sequences if obstacles/intersections ahead.
- If you only see sky and road on one side, it means you are at the map boundary. Rotate to face buildings.
- The subtask is done only when the current view matches the expected view.

Remember:
- If you believe the subtask is completed, output [-1]. Remind yourself in the history that you are starting to do the next subtask.
- You must always output at least one action. If lost, try rotating.

Output Format:
You must ALWAYS return a JSON exactly like:
{"Reason": "Your reason", "Summary": "New summary of history and current status", "Actions": [list of integers]}
\end{promptbox}

\vspace{-5pt}

\paragraph{Finetuning} \label{sec:app_qwen_finetune}

\begin{table}[H]
\centering
\small
\caption{Key hyperparameters used during finetuning}
\begin{tabular}{ll}
\toprule
\textbf{Hyperparameter} & \textbf{Value} \\
\midrule
Training epochs & 2 \\
Batch size & 32 (4$\times$8) \\
Optimizer & AdamW \\
Learning rate & 2e-4 \\
Weight decay & 0.01 \\
LR Scheduler & Constant (with warmup) \\
Warmup ratio & 0.03 \\
Gradient Clipping & 0.3 \\
LoRA rank & 8 \\
LoRA alpha & 16 \\
LoRA dropout & 0.05 \\
\bottomrule
\end{tabular}
\label{tab:finetune-hyper}
\end{table}

During finetuning, the model receives as input the current observation, the target image representing the expected view upon subtask completion, the instruction, the current orientation, and the ground-truth action history. Based on these inputs, the model is trained to predict the estimated distance, the anticipated final orientation, and the remaining action sequence as a form of CoT planning, followed by the prediction of the next action. We finetune the Qwen2.5-VL-7B-Instruct model using LoRA \cite{dettmers2023qloraefficientfinetuningquantized} applied to both the language model head and the merging projection layer. The loss is computed solely on the tokens generated by the decoder. We used four A100 GPUs with 80GB VRAM each for finetuning. The hyperparameters can be found in Table~\ref{tab:finetune-hyper}

\paragraph{Hybrid Method}
The hybrid system uses GPT-4o as a high-level decision maker determining whether to continue straight or turn at intersections. It then uses A* as a low-level path planner to generate and execute movement commands.
\paragraph{RL Method} The policy is first pretrained through behavioral cloning on expert demonstrations and then finetuned with PPO. Training is conducted on two NVIDIA L40S GPUs, each running two parallel instances of SWR.

\begin{promptbox}{System Prompts in Finetuned Single Agent Instruction Following}
You are a navigation robot in a 3D environment.

You will be given:
- An current egocentric image.
- An expected view image (what you should see when the subtask is completed).
- A textual description of the subtask.
- Your current cardinal direction (e.g., "North").
- A history of previously taken actions.

You must:
1. Determine whether the current subtask is already completed by comparing the current and expected views.
2. Deduce the expected orientation when the subtask is completed.
3. Deduce the distance from the current position to the expected position.
4. Plan the remaining actions to complete the subtask based on current and expected views.
5. If the subtask is not completed, output the action you will take in this step. If the subtask is completed, output -1.

Valid actions:
-1: Subtask_completed
0: Move_forward - Move 5 meters forward in the direction the robot is facing.
1: Rotate_left - Rotate 90° to the left.
2: Rotate_right - Rotate 90° to the right.

Output Format:
Only return a JSON object like:
{"Expected_Orientation": "The Orientation", "Remaining_Distance": "The Distance", "Remaining_Actions": "Textual Plan of the Actions", "Next_Action": integer}
\end{promptbox}

\subsection{More Quantitative Results}

\paragraph{Ablation} \label{sec:app_MMNav_quant}
Here we provide the ablation study of our ReAct baseline with GPT-4o as base model, tested on a 50-task subset of SimWorld-MMNav, as Table \ref{tab:ablation} shows.

\begin{table}[h!]
\centering
\small
\begin{threeparttable}
\caption{Ablation study with key components.}
\label{tab:ablation}
\begin{tabular}{@{}l
                >{\centering\arraybackslash}p{1.3cm}
                >{\centering\arraybackslash}p{1.6cm}
                >{\centering\arraybackslash}p{1.0cm}
                >{\centering\arraybackslash}p{1.5cm}
                >{\centering\arraybackslash}p{1.2cm}
                >{\centering\arraybackslash}p{2.4cm}@{}}
\toprule
\textbf{Configuration} & \textbf{Explicit Reason} & \textbf{Separate Perceive/Act} & \textbf{Depth} & \textbf{Segment} & \textbf{Strip} & \textbf{Subtask Success Rate (\%)} \\
\midrule
Our setting & \checkmark & \checkmark & -- & \checkmark & -- & 34.38 \\
Merged call & \checkmark & -- & -- & \checkmark & -- & 33.54 \\
w/ depth & \checkmark & \checkmark & \checkmark & \checkmark & -- & 33.39 \\
w/o explicit ReAct & -- & \checkmark & -- & \checkmark & -- & 32.21 \\
w/o segmentation & \checkmark & \checkmark & -- & -- & -- & 31.90 \\
w/ stripping & \checkmark & \checkmark & -- & \checkmark & \checkmark & 31.22 \\
\bottomrule
\end{tabular}
\end{threeparttable}
\vspace{-8pt}
\end{table}

Our setting requires the model to explicitly reason before acting through multi-turn interaction: it first describes the observation, then reasons and decides the action. The input includes the observation and its segmentation mask, without depth or stripped visual parts.

The perception-action framework simplifies the mapping process and reduces hallucinations.
While the model possesses a certain implicit depth estimation capacity, directly adding depth images yields marginal gain and can even introduce noise if the colormap is misaligned.
The explicit ReAct framework notably stabilizes the reasoning process and mitigates hallucinations in complex intersections.
Although GPT-4o itself lacks sufficient training on intersection-heavy scenes, the ground-truth segmentation image helps alleviate this limitation.
Finally, 
stripping the observation into vertical chunks increases visual matching difficulty, leading to degraded performance in our simplified setting. 

\paragraph{Statistical Significance}
We also provide the 95\% CI of our main results, as Table \ref{tab:single-agent-benchmark-CI} shows. For success rate and subtask success rate, we use binomial proportion confidence interval~\cite{wilson-ci} and display both lower and upper bound in the table.

\begin{table}[!h]
    \centering
    \caption{Experimental results on the SimWorld-MMNav benchmark (easy task set) with confidence intervals}
    \scriptsize
    \setlength{\tabcolsep}{16pt}
    \label{tab:single-agent-benchmark-CI}
    \resizebox{\textwidth}{!}{%
    \begin{tabular}{lccc}
    \toprule
    \textbf{Models} & \textbf{SR\%$\uparrow$} & \textbf{Subtask SR\% $\uparrow$} & \textbf{Distance Progress\% $\uparrow$} \\
    \midrule
    \rowcolor{gray!10} \multicolumn{4}{l}{\textit{Proprietary Models}} \\
    GPT-4o & 0 [0, 3.85] & 33.07 [24.71, 43.24] & 15.60 ($\pm$6.97) \\
    Gemini 2.5 Flash & 0 [0, 4.32] & 37.06 [28.09, 48.27] & 31.29 ($\pm$8.11) \\
    \midrule
    \rowcolor{gray!10} \multicolumn{4}{l}{\textit{Open-sourced Models}} \\
    QwenVL 2.5 7B & 0 [0, 4.14] & 16.86 [10.49, 25.96] & 7.82 ($\pm$2.45)\\
    QwenVL 2.5 72B & 0 [0, 3.89] & 23.80 [17.60, 34.84] & 17.50 ($\pm$6.11) \\
    Gemma 3 27B & 0 [0, 3.85] & 15.36 [9.70, 24.19] & 6.83 ($\pm$5.44) \\
    InternVL 3 78B & 0 [0, 4.28] & 18.31 [11.79, 28.11] & 9.34 ($\pm$4.04) \\
    \midrule
    \rowcolor{gray!10} \multicolumn{4}{l}{\textit{Fine-tuned Models}} \\
    QwenVL2.5 7B$_{ft}$  & 4.0 [1.08, 13.22] & 52.45 [39.52, 65.95] & 53.63 ($\pm$10.88) \\
    \bottomrule
    \end{tabular}
    }
    \vspace{-5pt}
\end{table}
\section{\textsc{SIMWORLD-MRS}}\label{sec:app_MRS}
To address the limitations of existing benchmarks in multi-robot search, we propose \textbf{SimWorld-MRS}, a new benchmark designed to evaluate collaboration, localization, and communication among multiple robots in large-scale, photo-realistic urban environments. SimWorld-MRS simulates realistic challenges such as partial observability, dynamic environments, and natural language coordination. In the following subsections, we detail the procedural task generation, baseline implementations, prompting strategies, and provide case studies to illustrate key behaviors like localization and communication.

\subsection{Detailed Task Settings}

\paragraph{Observation Space} The observation space for the follower robot closely mirrors that of the single-agent setting, with the key difference being that its instructions are exclusively derived from inter-agent communication. In contrast, the guide robot has access to the complete map and landmark information of the simulated city, enabling it to generate an oracle path plan and identify specific landmarks for rendezvous.

\paragraph{Action Space} Compared to the single-agent setting, the follower robot is additionally capable of initiating communication, while the guide robot is responsible for route planning and conveying instructions to the follower. Both robots are also able to pause and check whether the other is within their field of view, facilitating coordinated rendezvous.

\subsection{Metric Detail}

We report two metrics for evaluating performance on the SimWorld-MRS task: Collaborative Success Rate (CSR) and Task Progress (TP).

\paragraph{Collaborative Success Rate}  
This metric estimates the probability that the two robots successfully meet up across different maps. A meet-up is considered successful if at least one robot executes the \texttt{check\_task\_complete} action and detects the other robot dog in its observation via ground-truth segmentation.

\paragraph{Task Progress}  
This metric quantifies the relative reduction in distance between the two robots by the end of the task. Let $D_0$ denote the initial distance between the two agents, and $D_T$ the distance when the task terminates. The Task Progress (TP) is defined as:
\begin{equation}
\text{TP} = \max\left(\frac{D_0 - D_T}{D_0}, 0\right)
\label{eq:task_progress}
\end{equation}

\subsection{Procedural Task Generation}

To construct the SimWorld-MRS benchmark, we procedurally generate 100 unique city maps, each covering a large-scale urban environment. For each map, we instantiate one multi-robot search task, resulting in a total of 100 evaluation tasks. The task generation process follows Algorithm~\ref{alg:SimWorld-mrs-task-gen}.

\begin{algorithm}[ht]
\caption{Task Generation for SimWorld-MRS}
\KwIn{City map with $m$ streets; sample $n$ landmark buildings per street}
\KwOut{main robot's memory $\mathcal{M}$, main robot spawn $s_{\text{main}}$, following robot spawn $s_{\text{follow}}$}

\tcp{Phase 1: Build Main Robot’s Memory}
Initialize memory set $\mathcal{M} \leftarrow [\,]$\;

\For{$i \leftarrow 1$ \KwTo $m$}{
    Sample $n$ landmark buildings on street $i$: $\{L_{i1}, L_{i2}, \dots, L_{in}\}$\;

    \For{$j \leftarrow 1$ \KwTo $n$}{
        $x_{ij} \leftarrow$ get front-door location of $L_{ij}$\;

        $o_{ij} \leftarrow$ compute orientation facing toward $L_{ij}$\;

        $I_{ij} \leftarrow$ capture visual cue at $x_{ij}$ with orientation $o_{ij}$\;

        Add $(I_{ij}, x_{ij})$ to memory set $\mathcal{M}$\;
    }
}

\tcp{Phase 2: Sample Initial Robot Positions}
Obtain a set of valid robot spawn points $S$\;

Randomly sample two distinct spawn points from $S$: $s_{\text{main}}, s_{\text{follow}}$\;

\Return $(\mathcal{M}, s_{\text{main}}, s_{\text{follow}})$
\label{alg:SimWorld-mrs-task-gen}
\end{algorithm}

Specifically, we first sample $n$ landmark buildings along each of the $m$ streets in the city and collect their front-door images with aligned orientations. These image-location pairs serve as the main robot's memory of the city. Next, we sample two distinct and valid spawn locations as the starting positions for the main robot and the following robot. This ensures spatial diversity and supports realistic localization and communication challenges.

To evaluate the accuracy of the final meetup, we utilize the observations captured by both robots at the end of the multi-agent navigation process. Given that our system includes access to ground-truth segmentation, we determine whether the other robot appears within the field of view by inspecting the segmentation output. The presence of the counterpart robot in the observation is treated as evidence of a successful meetup.

\subsection{Baseline Detail}\label{sec:app_MRS_Baselines}
\paragraph{Baseline 1 - Oracle Planner}

In the multi-robot oracle setting, our baseline communication pipeline consists of three key components: the follower robot's description of the current building, the main robot's retrieval of the building, and the oracle path planning to reach the follower.

The description module takes the current egocentric observation as input and feeds it into a VLM using a templated prompt. This prompt is specifically designed to guide the model towards generating goal-oriented and informative descriptions of the current scene, ensuring alignment with task requirements.

The memory retrieval module receives the generated textual description and a pre-collected landmark image dataset as input. It then uses the VLM to identify the landmark image that best matches the description. To achieve this, we implement a tournament-style elimination process: images are compared in pairs based on their semantic and visual alignment with the description, and in each round, the less relevant image is discarded. This process continues iteratively until a single most relevant image remains. To mitigate hallucinations and enhance retrieval stability, we employ a system prompt that encourages precise and comparative reasoning. The output of this module is the position of the landmark image judged by the VLM to be the most semantically and visually consistent with the description.

The oracle path planning module receives as input the current location of the main robot and the estimated location of the follower robot, as determined by the memory retrieval module. Using full access to the global city map, the module computes the shortest collision-free path between the two locations via the A* algorithm. The resulting path is converted into a sequence of discrete navigation actions, such as moving forward or turning at intersections. These actions serve as an oracle reference and can be directly issued to the follower robot or translated into natural language instructions for communication purposes.

During the main robot’s execution of the oracle path, the follower robot actively rotates at each step and uses the VLM to detect whether the other robot dog appears in its field of view.

\begin{figure}[!h]
    \centering
    \includegraphics[angle=270, trim=80 0 80 0, clip,width=\linewidth]{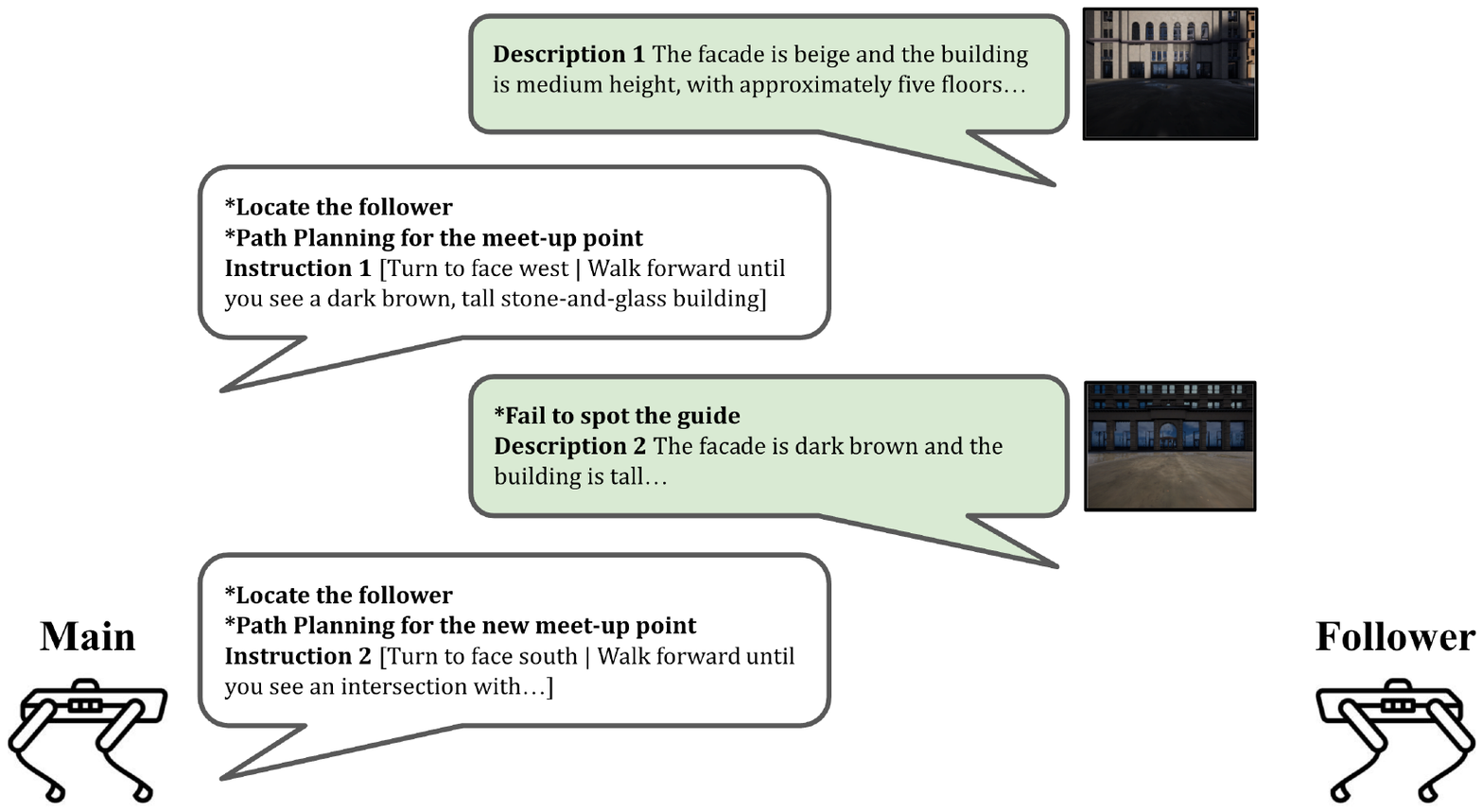}
    \caption{Example communication for ROCO baseline}
    \label{fig:roco}
\end{figure}

\paragraph{Baseline 2 - ROCO}

The ROCO-based \cite{mandi2023rocodialecticmultirobotcollaboration} setting extends the oracle setup by introducing collaborative planning and communication between two robots. After the two agents communicate and confirm the follower’s current location in terms of a landmark building, the main robot uses access to the ground-truth map to identify the landmark closest to the midpoint between both agents. It then performs path planning to compute the optimal trajectories for both agents to converge at this intermediate landmark. The main robot transforms the computed path for the follower into natural language instructions, which are communicated to the follower robot. The follower then performs multimodal instruction following to navigate toward the meeting point and rendezvous with the main robot. An example for communication in the ROCO baseline can be seen in Figure \ref{fig:roco}.

The description, memory retrieval, and path planning modules mirror those in the oracle baseline. The instruction generation module translates the planned path into natural language commands using a predefined set of sentence templates. These instructions are compact and landmark-aware, designed to guide the follower robot step-by-step without access to the full map. Each instruction encapsulates one or more discrete navigation actions, such as moving forward until reaching a visible landmark or turning at an intersection.

The communication module governs when to initiate a new dialogue round. After executing a received instruction, the follower robot monitors its state and triggers a new communication cycle under the condition: if it believes it has completed the instruction but, after rotating to search, does not observe the main robot. Upon reactivation, the follower generates a new scene description, enabling the main robot to re-localize its position and update the rendezvous plan accordingly.

The instruction-following module remains largely consistent with the single-agent setting, with the primary difference being that the visual hint is no longer provided.

\paragraph{Prompting for Multi-agent Baselines}

We adopt a modular prompting strategy that aligns with our multi-agent architecture. Specifically, we design two system prompts—one for generating egocentric scene descriptions (used by the follower robot), and another for conducting comparative landmark retrieval (used by the main robot). These prompts are implemented via a VLM with image and text modality support.

The follower robot observes its local environment and produces a textual description aimed at helping the main robot identify its current location. The prompt guides the VLM to emphasize salient and matchable visual features such as color, height, materials, signage, and local context.

\begin{promptbox}{Follower Description Prompt}
You are an expert building-description assistant.

In ≤200 words, describe the building so another person could match photos of it.

**Start the first sentence with the facade's MAIN COLOR and HEIGHT.**

Cover these attributes as comma-separated phrases:
- main color — dominant facade color
- height — low / medium / tall (number of floors)
- primary materials — e.g., brick, concrete, glass, steel
- window grid / pattern — shape and arrangement of windows
- ground-floor layout — doors, arches, glazing style
- signage text — exact words visible; say "no signage" if none
- sidewalk objects — lamp-posts, trees, benches, etc.
- distinctive features — murals, balconies, arches, etc.
- neighbor — immediate surrounding context (e.g., adjacent buildings or empty lots)

Keep it factual and avoid subjective opinions. Return a single descriptive paragraph in natural language.
\end{promptbox}

After receiving a description from the follower robot, the main robot attempts to match it against its landmark memory using a VLM-based comparative prompt. The model is required to select the image (A or B) that better matches the textual input, focusing on distinctive visual features.

\begin{promptbox}{Main Robot Retrieval Prompt}
TEXT = Natural-language description of the target building.

Two FULL-FACADE candidate photos are shown: A (first) and B (second).

Decide which matches TEXT better.

Guidelines:

- Give strong weight to facade color, signs, materials, and distinctive features.
- If the facade color clearly mismatches, that candidate must lose.
- You must compare both options and select the better one.

Reply only with A or B.
\end{promptbox}
\section{Qualitative Examples}\label{sec:app_QE}

We provide qualitative examples to illustrate the common failure modes of VLMs.

\vspace*{-5pt}
\subsection{SimWorld-MMNav}\label{sec:app_QE_MMNav}
\vspace*{-5pt}

\begin{figure}[!h]
    \centering
    \includegraphics[angle=270, trim=50 0 50 0, clip,width=\linewidth]{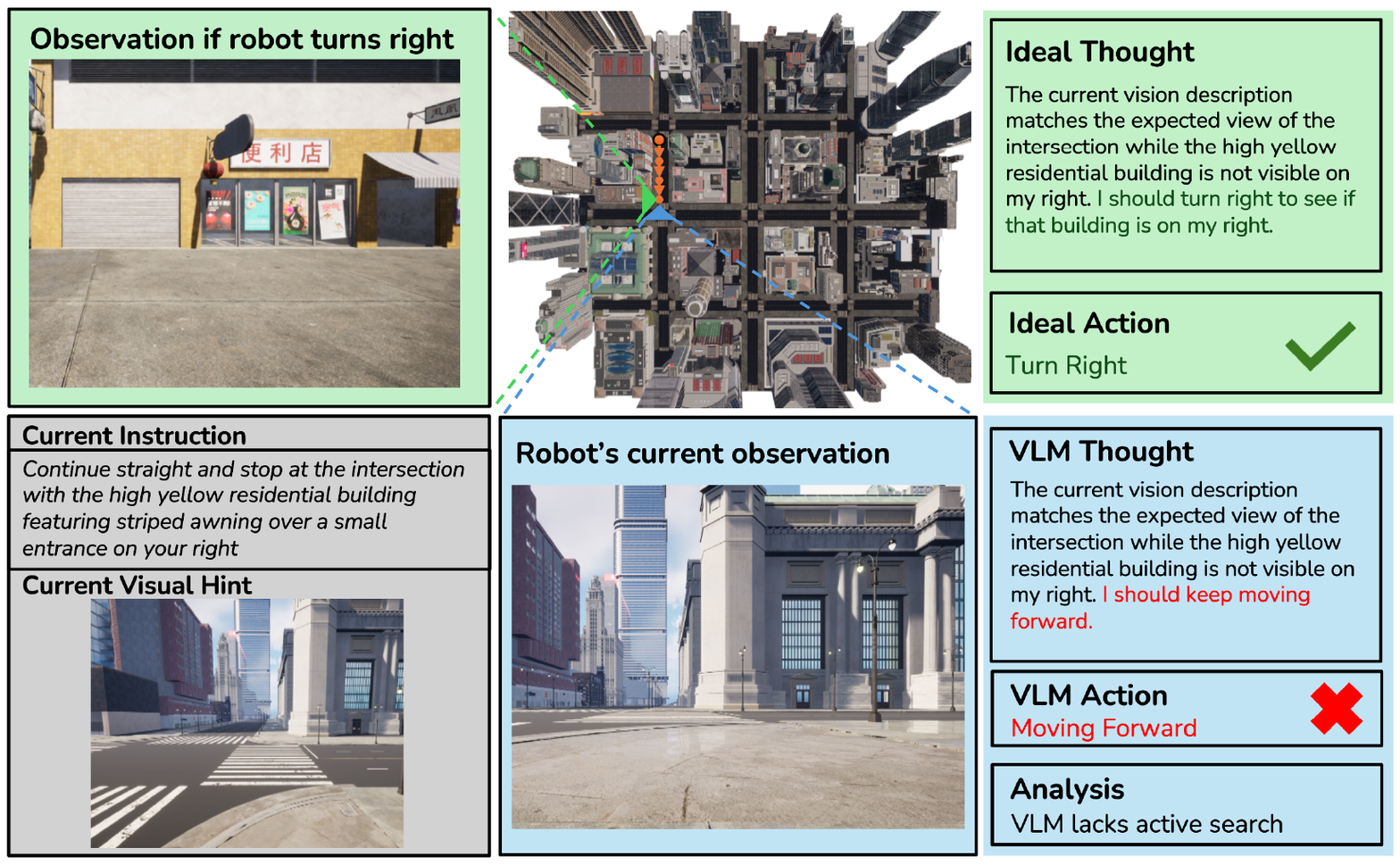}
    \caption{
    Qualitative result - lack of active perception}
    \label{fig:case 1}
\end{figure}

\paragraph{Active Perception} The VLM lacks initiative in active perception. As Figure \ref{fig:case 1} shows, at intersections, the robot’s field of view is often limited, and the landmark referenced in the instruction may not be visible from the current perspective. As a result, the robot does not recognize the location as the intended intersection, even when the visual hint has already been matched. Ideally, the robot should rotate to check its surroundings for the missing landmark. However, the VLM tends to continue moving forward, waiting for the landmark to appear directly in front of the robot rather than actively seeking it through lateral exploration.

\begin{figure}[!h]
    \centering
    \includegraphics[angle=270, trim=190 5 190 10, clip,width=\linewidth]{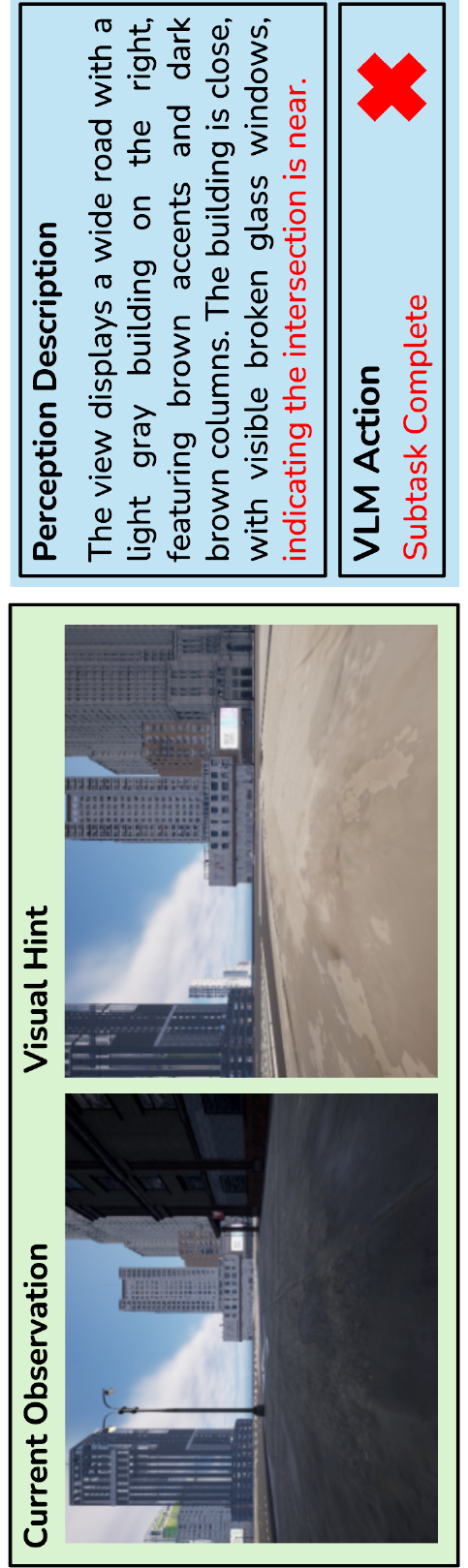}
    \vspace*{-5pt}
    \caption{Qualitative result - lack of distance grounding}
    \label{fig:case 2}
\end{figure}

\paragraph{Spatial Reasoning} The VLM exhibits limitations in reasoning about spatial relationships, particularly in estimating distance, maintaining spatial continuity, and interpreting alternate perspectives. In one failure case, the current observation partially resembles the visual hint, leading the VLM to prematurely assume arrival at the intersection, despite the robot still being far from the crosswalk, as Figure \ref{fig:case 2} shows. This example also indicates that the recognition of intersection lacks robustness and relies heavily on referential landmark buildings.

\begin{figure}[!h]
    \centering
    \includegraphics[angle=270, trim=160 8 160 10, clip,width=\linewidth]{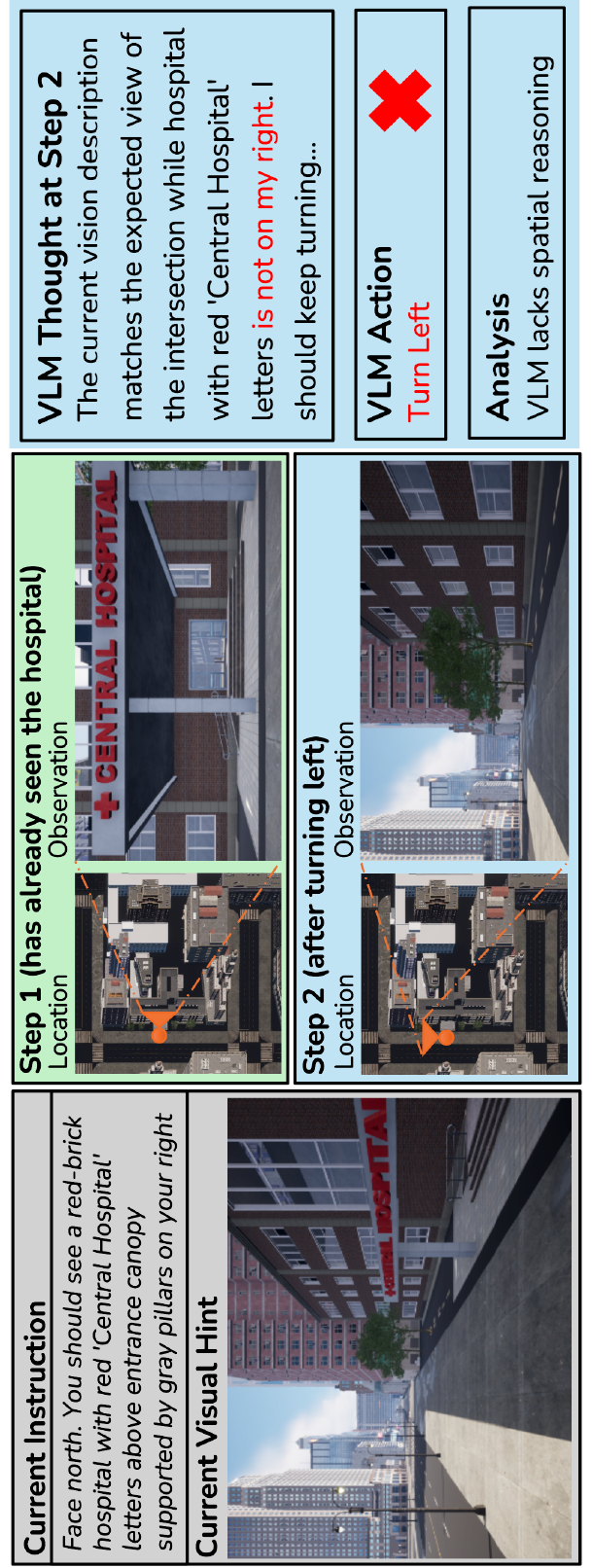}
    \vspace*{-5pt}
    \caption{Qualitative result - lack of embodied reasoning}
    \label{fig:case 3}
\end{figure}

In another instance, as Figure \ref{fig:case 3} shows, during an orientation alignment task, the robot is initially facing a landmark. After turning right, a characteristic part of the landmark disappears from view. Given a working memory, an embodied agent would robustly infer that it has aligned accordingly. However, the VLM fails to make this inference, indicating a lack of embodied spatial understanding.

\begin{figure}[!h]
    \centering
    \includegraphics[angle=270, trim=190 5 190 10, clip,width=\linewidth]{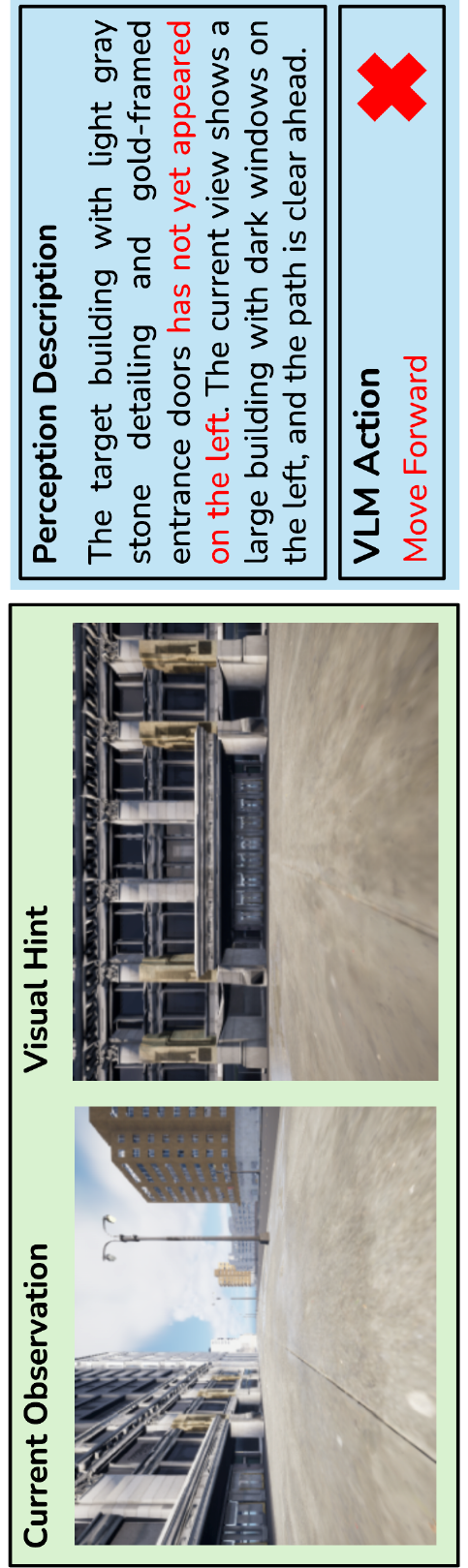}
    \vspace*{-5pt}
    \caption{Qualitative result - lack of perspective-adaptive matching}
    \label{fig:case 4}
\end{figure}
\vspace*{-5pt}

These limitations also manifest when matching buildings from different perspectives. The target building is provided as a frontal image, but during navigation, only a side view may be visible. The perception module often fails to associate the side and frontal views and provides insufficient information, causing the robot to overlook the destination, as Figure \ref{fig:case 4} shows. 

Even when a correct match is made, the VLM may still fail to reorient. For example, given the thought, "Currently, I am on a wide street facing east with the high brown concrete building and billboard on my left. The billboard is not directly in front of me yet, indicating that I need to continue moving forward until it is," the model chooses to proceed rather than rotate, failing to reason from an embodied perspective.

\paragraph{Pragmatic Reasoning} The VLM also struggles with interpreting pragmatic intent in natural instructions. When given the instruction "turn right at the intersection," it often treats the subtask as complete after a single turning action. In real-world scenarios, such a turn typically involves multiple steps, like moving forward, turning, and possibly crossing the street. The VLM’s overly literal interpretation leads to partial execution and deviation from the intended path.

\vspace*{-5pt}
\subsection{A Success Case of Finetuned Baseline}
\vspace*{-5pt}

\begin{figure}[!h]
    \centering
    \includegraphics[angle=270, trim=110 0 110 0, clip, width=\linewidth]{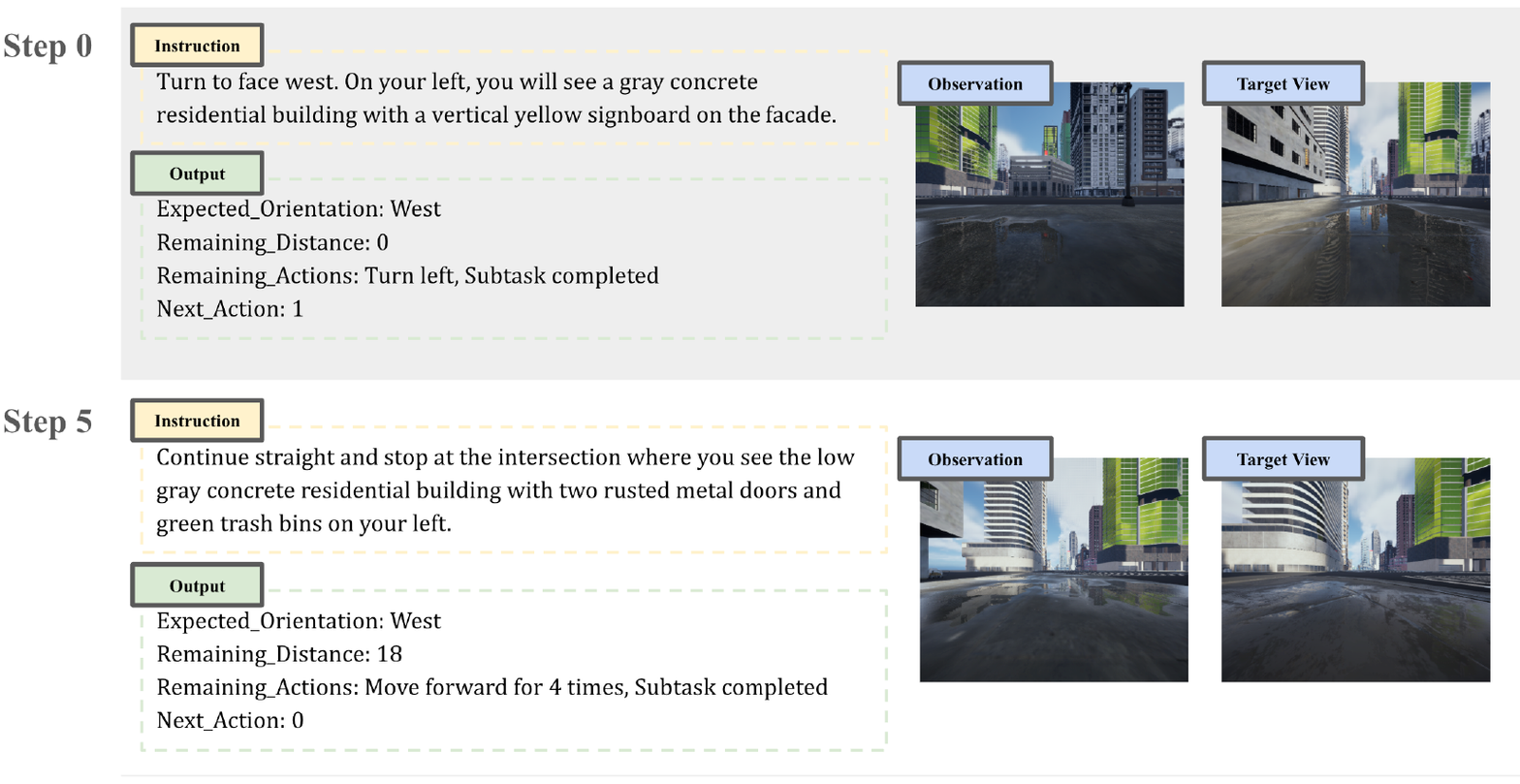}
    \vspace{0pt}
    \includegraphics[angle=270, trim=120 0 120 0, clip, width=\linewidth]{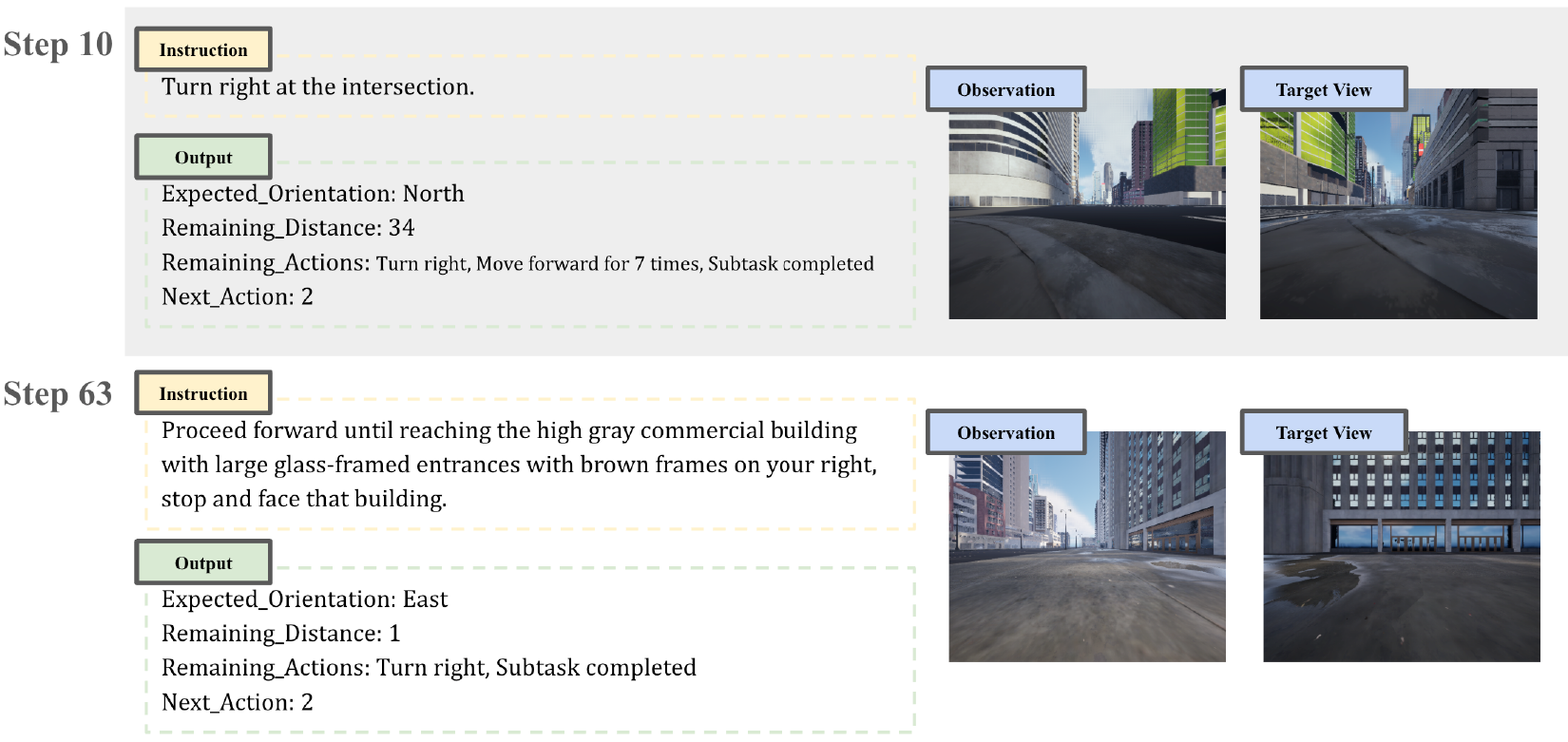}
    \caption{Qualitative result – key-step VLM outputs from the finetuned model successfully completing the task}
    \label{fig:finetune-success}
\end{figure}

As Figure~\ref{fig:finetune-success} shows, finetuning facilitates Qwen2.5-VL-7B-Instruct's spatial reasoning capabilities. When the agent is relatively close to the subtask target, the model is able to accurately infer the remaining distance and the corresponding sequence of actions, while also correctly predicting the final orientation. This enables a strong foundation for chain-of-thought (CoT) planning and improves the reliability of next-step action prediction. Furthermore, in turning scenarios, the finetuned model demonstrates, for the first time, the ability to reason about intermediate steps—such as the need to cross the street—in order to align with the target image. This allows the model to successfully complete the illustrated navigation task.

However, finetuning also exhibits certain limitations. First, when the target image corresponds to a location far from the agent's current observation, the lack of meaningful visual overlap makes it difficult for the model to reason about progress. Second, the model is finetuned solely on ground-truth action sequences and lacks robustness to error correction. As a result, if a mistake occurs during navigation, the model struggles to recover, which partially explains the still limited overall success rate.

\vspace*{-5pt}
\subsection{SimWorld-MRS}\label{sec:app_QE_MRS}
\vspace*{-5pt}

\paragraph{Salient Feature Ignorance}  In several failure cases, the follower robot generates descriptions that miss highly distinctive features such as store signs, murals, or logos. Instead, the VLM focuses on general elements like “a modern building with glass windows,” which are insufficient for precise localization. This results in ambiguous matches and large localization errors during memory retrieval by the main robot.

\paragraph{Failure of Instruction Following} The issues observed in single-agent instruction following often persist in this setting and are further exacerbated by the absence of a visual hint, making accurate instruction execution more challenging. Although repeated communication can partially correct navigation drift, the task may still fail if the follower agent stops at a non-landmark building, as the main robot will be unable to localize it for subsequent rendezvous planning.

\paragraph{Lack of Failure-Awareness}  While executing instructions, the follower robot often struggles to determine whether it has become lost. Even when the subtask has not been completed, and the action history indicates that the robot has already moved forward for an extended sequence, the model tends to continue moving forward until the task termination conditions are met. As a result, the robot fails to recognize its deviation in time to trigger communication for goal correction, ultimately leading to out-of-bounds behavior or exceeding the maximum allowed number of steps.

\section{Code Availability Statement}\label{sec:open_source}

The implementation of our system builds on a codebase developed through multi-institutional collaboration, with components that are difficult to anonymize due to dependency structures and prior repository history. To maintain the integrity of the double-blind review process, we have withheld the release of the code at this stage. We are committed to open science and will publicly release the complete codebase, along with detailed documentation and instructions for reproduction, upon acceptance.

\section{LLM Usage Statement}\label{sec:LLM_usage}

We utilize several large multimodal models (VLMs) as core components of our study, including GPT-4o, Gemini 2.5 Flash, Gemini 2.0 Flash, Qwen2.5-VL-72B-Instruct, Qwen2.5-VL-7B-Instruct, InternVL-78B, and Gemma3-27b-it. These models are evaluated within our simulator-based benchmarks to investigate their embodied navigation capabilities and multi-agent communication performance. The LLMs are responsible for interpreting visual-linguistic instructions, reasoning about spatial environments, and generating actions or dialogue.

Furthermore, we fine-tune Qwen2.5-VL-7B-Instruct on our proposed training dataset to assess the effectiveness of task-specific supervision. Since the models play a central role in both methodology and experimental analysis, and significantly influence the reported results, we declare their usage as integral to the core of this research.

\clearpage
\newpage
\section*{NeurIPS Paper Checklist}

\begin{enumerate}

\item {\bf Claims}
    \item[] Question: Do the main claims made in the abstract and introduction accurately reflect the paper's contributions and scope?
    \item[] Answer: \answerYes{} 
    \item[] Justification: The abstract summarizes the key features of our work, focusing on what differentiates our simulator from existing ones—the main focus of this paper.

\item {\bf Limitations}
    \item[] Question: Does the paper discuss the limitations of the work performed by the authors?
    \item[] Answer: \answerYes{} 
    \item[] Justification: We discussed limitations in the conclusion.

\item {\bf Theory assumptions and proofs}
    \item[] Question: For each theoretical result, does the paper provide the full set of assumptions and a complete (and correct) proof?
    \item[] Answer: \answerNA{} 
    \item[] Justification: The paper does not include theoretical results.

    \item {\bf Experimental result reproducibility}
    \item[] Question: Does the paper fully disclose all the information needed to reproduce the main experimental results of the paper to the extent that it affects the main claims and/or conclusions of the paper (regardless of whether the code and data are provided or not)?
    \item[] Answer: \answerYes{} 
    \item[] Justification: We will open-source our gym environment, the agent's prompt, backend, and model. Please check the details of our two benchmarks and the baselines implementation in Appendix~\ref{sec:app_MMNAv} and Appendix~\ref{sec:app_MRS}

\item {\bf Open access to data and code}
    \item[] Question: Does the paper provide open access to the data and code, with sufficient instructions to faithfully reproduce the main experimental results, as described in supplemental material?
    \item[] Answer: \answerYes{} 
    \item[] Justification: We will open-source our code as mentioned in Appendix~\ref{sec:open_source}and the details of baseline implementation can be found in Appendix~\ref{sec:app_MMNAv_baselines} and Appendix~\ref{sec:app_MRS_Baselines}.

\item {\bf Experimental setting/details}
    \item[] Question: Does the paper specify all the training and test details (e.g., data splits, hyperparameters, how they were chosen, type of optimizer, etc.) necessary to understand the results?
    \item[] Answer: \answerYes{} 
    \item[] Justification: We have specified that 33 percent of buildings in the training set are excluded from testing environment.

\item {\bf Experiment statistical significance}
    \item[] Question: Does the paper report error bars suitably and correctly defined or other appropriate information about the statistical significance of the experiments?
    \item[] Answer: \answerYes{} 
    \item[] Justification: Each experiment is run twice to account for variability, and results are reported accordingly. While limited in sample size, this provides an initial estimate of consistency. 

\item {\bf Experiments compute resources}
    \item[] Question: For each experiment, does the paper provide sufficient information on the computer resources (type of compute workers, memory, time of execution) needed to reproduce the experiments?
    \item[] Answer: \answerYes{} 
    \item[] Justification: We use on a headless machine with an AMD EPYC 9534 CPU, L40S GPU, 64GB RAM
    
\item {\bf Code of ethics}
    \item[] Question: Does the research conducted in the paper conform, in every respect, with the NeurIPS Code of Ethics \url{https://neurips.cc/public/EthicsGuidelines}?
    \item[] Answer: \answerYes{} 
    \item[] Justification: We follow the NeurIPS Code of Ethics.

\item {\bf Broader impacts}
    \item[] Question: Does the paper discuss both potential positive societal impacts and negative societal impacts of the work performed?
    \item[] Answer: \answerNA{} 
    \item[] Justification: While we do not explicitly discuss societal impacts, we believe our simulator and benchmark can positively contribute to research in embodied AI, and we do not foresee any negative societal consequences from this work.
    
\item {\bf Safeguards}
    \item[] Question: Does the paper describe safeguards that have been put in place for responsible release of data or models that have a high risk for misuse (e.g., pretrained language models, image generators, or scraped datasets)?
    \item[] Answer: \answerNA{} 
    \item[] Justification: Our work has no misuse risk.

\item {\bf Licenses for existing assets}
    \item[] Question: Are the creators or original owners of assets (e.g., code, data, models), used in the paper, properly credited and are the license and terms of use explicitly mentioned and properly respected?
    \item[] Answer: \answerYes{} 
    \item[] Justification: Yes. All assets used in our work were purchased from the official Unreal Engine Marketplace (fab.com), and we fully comply with their licensing terms. No assets were used for any unauthorized or additional 3D asset generation.

\item {\bf New assets}
    \item[] Question: Are new assets introduced in the paper well documented and is the documentation provided alongside the assets?
    \item[] Answer: \answerYes{} 
    \item[] Justification:  Yes. We introduce a 20K training dataset to support vision-language navigation tasks, and detail of it can be found in appendix~\ref{sec:app_20k}.

\item {\bf Crowdsourcing and research with human subjects}
    \item[] Question: For crowdsourcing experiments and research with human subjects, does the paper include the full text of instructions given to participants and screenshots, if applicable, as well as details about compensation (if any)? 
    \item[] Answer: \answerNA{} 
    \item[] Justification: We do not involve any research with human subjects.

\item {\bf Institutional review board (IRB) approvals or equivalent for research with human subjects}
    \item[] Question: Does the paper describe potential risks incurred by study participants, whether such risks were disclosed to the subjects, and whether Institutional Review Board (IRB) approvals (or an equivalent approval/review based on the requirements of your country or institution) were obtained?
    \item[] Answer: \answerNA{}{} 
    \item[] Justification: We do no involve crowdsourcing nor research with human subjects.

\item {\bf Declaration of LLM usage}
    \item[] Question: Does the paper describe the usage of LLMs if it is an important, original, or non-standard component of the core methods in this research? Note that if the LLM is used only for writing, editing, or formatting purposes and does not impact the core methodology, scientific rigorousness, or originality of the research, declaration is not required.
    \item[] Answer: \answerYes{} 
    \item[] Justification: We provide a detailed explanation in the experimental section on how large language models (LLMs) and vision-language models (VLMs) are used to conduct experiments, which form a core component in our benchmark. Details of it can be found in appendix~\ref{sec:LLM_usage}

\end{enumerate}
\end{document}